\documentclass[preprint,5p,times,twocolumn]{elsarticle}

\usepackage{amssymb}
\usepackage{amsmath}
\usepackage{array}
\usepackage{multirow}
\usepackage{booktabs} 
\usepackage{algorithm,algorithmic}
\usepackage{wrapfig} 
\usepackage{flushend} 
\usepackage{tabularx}
\usepackage{tikz}
\usetikzlibrary{positioning, fit, shapes, arrows}
\usepackage{multirow, makecell}
\usepackage{caption}
\usepackage{longtable}
\usepackage[hidelinks]{hyperref}
\usepackage{fontawesome}   
\newcolumntype{C}[1]{>{\centering\arraybackslash}m{#1}}

\journal{Information Fusion}

\begin{document}

\begin{frontmatter}




\title{UAVs Meet LLMs: Overviews and Perspectives Toward Agentic Low-Altitude Mobility}



\author[label1]{Yonglin Tian\fnref{equ1}}
\author[label2]{Fei Lin\fnref{equ1}}
\author[label2]{Yiduo Li}
\author[label2]{Tengchao Zhang}
\author[label3]{Qiyao Zhang}
\author[label2]{Xuan Fu}
\author[label2]{Jun Huang}
\author[label1]{Xingyuan Dai}
\author[label1]{\\Yutong Wang}
\author[label4]{Chunwei Tian}
\author[label5]{Bai Li}
\author[label1]{Yisheng Lv}
\author[label6]{Levente Kovács}
\author[label1,label2]{Fei-Yue Wang}

\fntext[cor1]{Equal contribution}
\affiliation[label1]{organization={The  State Key Laboratory of Multimodal Artificial Intelligence Systems,  Institute of Automation, Chinese Academy of Sciences},
            city={Beijing},
            postcode={100190}, 
            state={},
            country={China}}
            
\affiliation[label2]{organization={Department of Engineering Science, Faculty of Innovation Engineering, Macau University of Science and Technology},
            city={Macau},
            postcode={999078}, 
            state={},
            country={China}}
\affiliation[label3]{organization={School of Automation, Beijing Institute of Technology},
            city={Beijing},
            postcode={100081}, 
            state={},
            country={China}}           
\affiliation[label4]{organization={School of Software, Northwestern Polytechnical University},
            city={Xi'an},
            postcode={710129}, 
            state={},
            country={China}}
            
\affiliation[label5]{organization={College of Mechanical and Vehicle Engineering, Hunan University},
            city={Changsha},
            postcode={410082}, 
            state={},
            country={China}}
\affiliation[label6]{organization={John von Neumann Faculty of Informatics, Obuda University},
            city={Budapest},
            postcode={H-1034}, 
            state={},
            country={Hungary}}        
            

            
\begin{abstract}
Low-altitude mobility, exemplified by unmanned aerial vehicles (UAVs), has introduced transformative advancements across various domains, like transportation, logistics, and agriculture. Leveraging flexible perspectives and rapid maneuverability, UAVs extend traditional systems' perception and action capabilities, garnering widespread attention from academia and industry. However, current UAV operations primarily depend on human control, with only limited autonomy in simple scenarios, and lack the intelligence and adaptability needed for more complex environments and tasks. The emergence of large language models (LLMs) demonstrates remarkable problem-solving and generalization capabilities, offering a promising pathway for advancing UAV intelligence. This paper explores the integration of LLMs and UAVs, beginning with an overview of UAV systems' fundamental components and functionalities, followed by an overview of the state-of-the-art LLM technology. Subsequently, it systematically highlights the multimodal data resources available for UAVs, which provide critical support for training and evaluation. Furthermore, key tasks and application scenarios where UAVs and LLMs converge are categorized and analyzed.  Finally, a reference roadmap towards agentic UAVs is proposed to enable UAVs to achieve agentic intelligence through autonomous perception, memory, reasoning, and tool utilization. Related resources are available at \href{https://github.com/Hub-Tian/UAVs_Meet_LLMs}{https://github.com/Hub-Tian/UAVs\_Meet\_LLMs}.
\end{abstract}



\begin{keyword}


Unmanned aerial vehicles \sep large language models \sep foundation intelligence \sep low altitude mobility systems
\end{keyword}

\end{frontmatter}


\section{Introduction}

The rapid development of UAVs has introduced transformative solutions for monitoring and transportation across various sectors, including intelligent transportation, logistics, agriculture, and industrial inspection. With their flexible spatial mobility, UAVs significantly enhance the perception and decision-making capabilities of intelligent systems, offering a robust approach for upgrading traditional systems and improving operational efficiency. Given these advantages, UAV technology has attracted substantial attention from both academic researchers and industry practitioners.

Despite their promising potential, current UAV systems face several unique challenges due to characteristics that differentiate UAVs from ground-based or surface-based vehicles and robots:

\begin{itemize}
    \item \textbf{Flexible Viewing Angles:} The mobility of UAVs allows them to change their viewing angles dynamically, leading to variations in the perspectives from which objects are observed \cite{huang2022ufpmp, zhu2021tph}. This variability in input distribution introduces difficulties in tasks such as visual perception and reasoning.
    \item \textbf{Variable Altitudes:} UAVs operate at varying altitudes, which significantly impacts both the scale of observed objects and the field of view. These parameters fluctuate depending on the UAV’s flight height, making it challenging to maintain consistent environmental understanding during missions.
    \item \textbf{Three-Dimensional Mobility:} UAVs' ability to navigate in three-dimensional space adds complexity to tasks involving environmental perception and control. This requirement for 3D awareness elevates the difficulty of mission planning and execution, as it involves dynamic spatial reasoning and real-time adjustments to flight paths.
    \item \textbf{Swarming Requirements:} Compared to other autonomous agents, UAVs are increasingly required to operate in coordinated swarms. This need for collective action introduces challenges such as swarm interaction and behavior control, which require advanced algorithms for synchronization, communication, and task delegation.
    \item \textbf{Diverse Operational Environments:} UAVs are used in a wide range of applications, exposing them to a variety of heterogeneous environments. This diversity increases the complexity of object recognition and scene interpretation, as UAV systems must handle a wide array of objects and scenarios, often under unpredictable conditions \cite{yang2020advancing}.
\end{itemize}

These unique challenges underscore the need for advanced perception, reasoning, and coordination systems in UAV operations, particularly as they move from controlled environments to more complex, dynamic real-world applications. Currently, the majority of UAVs depend on human operators for flight control. This dependency not only incurs high labor costs but also introduces safety risks, as operators are limited by the range and sensitivity of onboard sensors when assessing environmental conditions. Such limitations impede the scalability and broader application of UAVs in complex environments. 

\begin{figure*}[ht]
 \centering
 \includegraphics[width=1\textwidth, trim=0cm 14.5cm 0cm 0cm, clip]{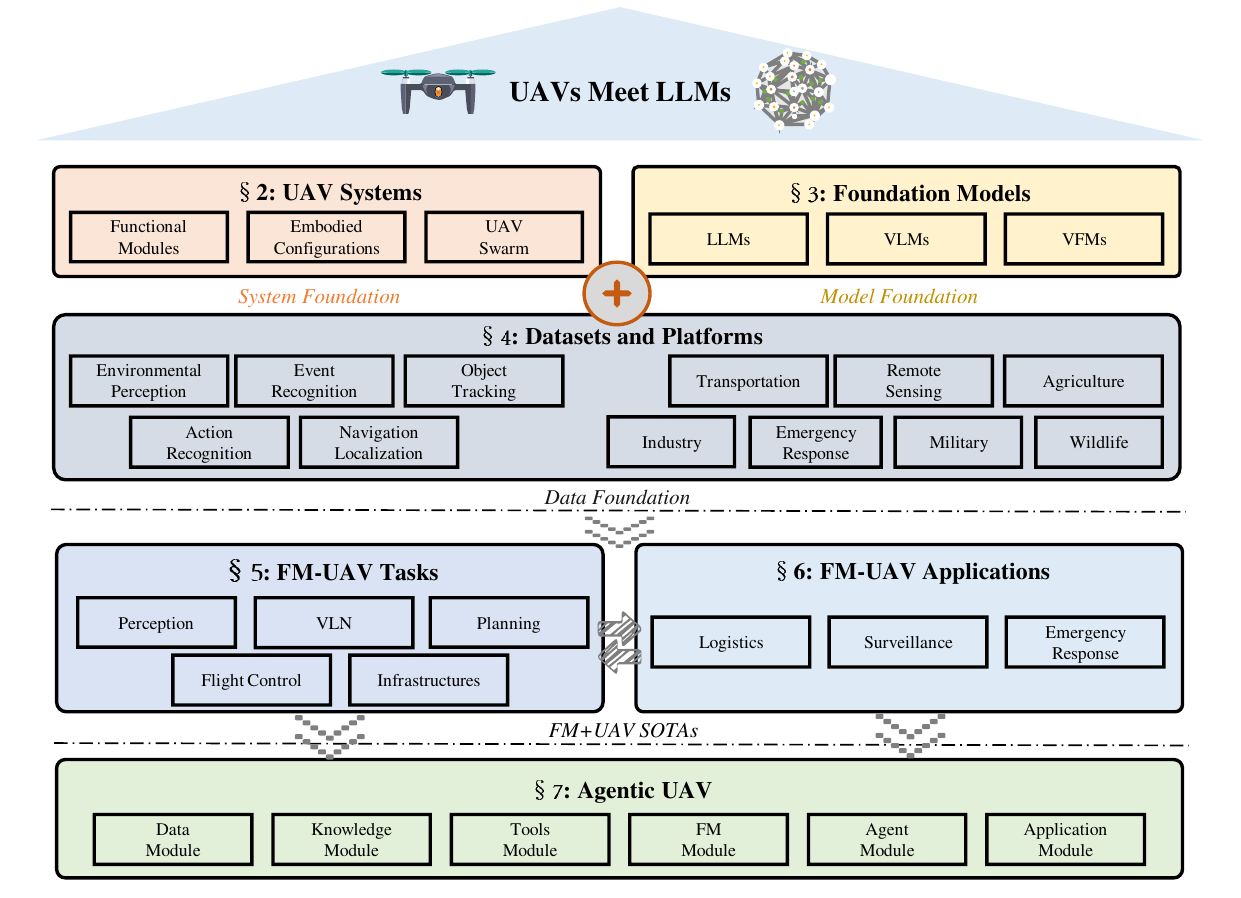}
 \caption{Main sections and the structure of this paper}
 \label{fig_roadmap}
\end{figure*}

Recent advancements in artificial intelligence, particularly in foundation models (FMs) such as ChatGPT \cite{vemprala2024chatgpt}, SORA, and various AI-generated content (AIGC) frameworks, have catalyzed significant transformations across industries \cite{wang2024parallel}. LLMs are endowed with near-human levels of commonsense reasoning and generalization capabilities, enabling advanced understanding, flexible adaptation, and real-time responsiveness in diverse applications. The integration of LLMs with UAV systems offers a promising avenue to enhance autonomy, providing UAVs with advanced reasoning capabilities and enabling more effective responses to dynamic environments.

Initial studies have explored integrating LLMs with UAVs in areas such as navigation \cite{liu2024navagent,sarkar2024gomaa}, perception\cite{liu2024shooting,qiu2024dronegpt}, planning \cite{chen2023typefly,tagliabue2023real}.  These early efforts highlight the potential of combining LLMs with UAV systems to foster more sophisticated autonomous behaviors. However, there remains a lack of systematic reviews on the integration of LLMs and UAVs, particularly regarding the frameworks and methodologies that support this interdisciplinary convergence. To advance the understanding of UAV and LLM integration, this paper provides a systematic review of the existing frameworks and methodologies, offering insights into the potential pathways for further advancing this interdisciplinary convergence. The main contributions of this paper are as follows.
\begin{itemize}
    \item A comprehensive background on the integration of UAVs and FMs is presented, detailing the fundamental components and functional modules of UAV systems while summarizing typical FMs. An extensive inventory of publicly available datasets is also provided, underscoring their crucial role in the development, training, and evaluation of intelligent UAV systems.
    \item A thorough review of recent studies on integrating LLMs with UAVs is conducted, highlighting essential methodologies, diverse applications, and the key challenges encountered in navigation, perception, and planning tasks.
    \item An agentic framework named Agentic UAVs is proposed, outlining the necessary architecture and capabilities to enable UAVs to achieve autonomous perception, reasoning, memory, and tool utilization, paving the way for their advancement into more intelligent and adaptable systems.

\end{itemize}

Through these contributions, we aim to provide a foundational overview of the current research landscape at the intersection of UAV technology and LLMs, highlight emerging trends and challenges, and propose directions for future investigation. This survey aspires to serve as a reference for researchers and practitioners seeking to leverage LLM capabilities to advance UAV autonomy and broaden the application potential of unmanned low-altitude mobility systems. The organization of this paper is illustrated in Figure \ref{fig_roadmap}. The system knowledge of UAVs and FMs is introduced from three perspectives: system foundation, model foundation, and data foundation. Subsequently, the integration of UAVs with FMs is explored, highlighting the state of the arts (SOTAs) in various tasks and applications. Finally, the architecture of agentic UAVs is proposed, outlining the objectives for future development.

\section{Systematic Overview of UAV Systems}

\begin{figure}[ht]
 \centering
 \includegraphics[scale=0.2]{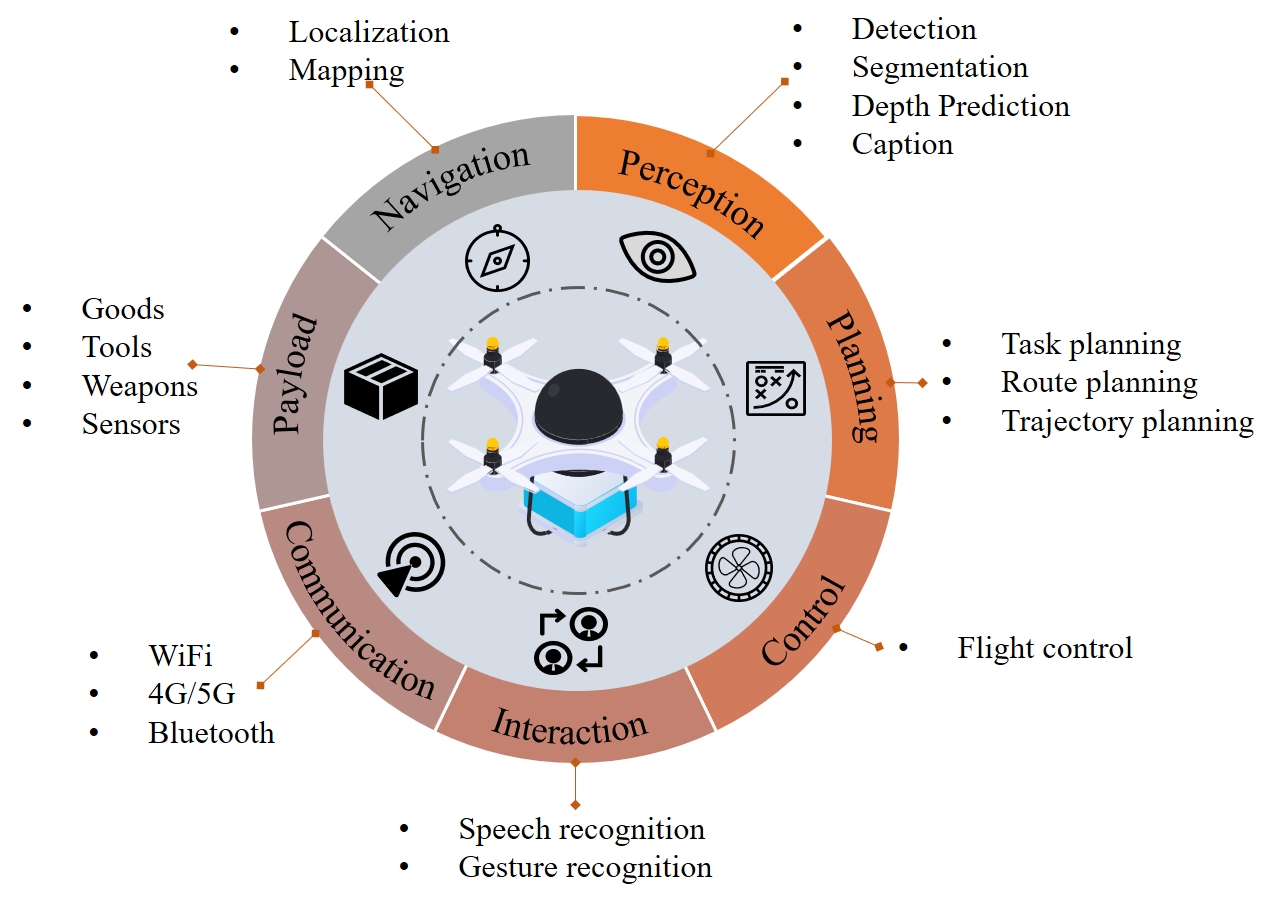}
 \caption{Key Functional modules of UAV systems}
 \label{fig_functions}
\end{figure}

This section provides a brief overview of intelligent UAVs from the perspectives of functional modules and embodied configurations. The functional modules encompass the core components of UAV systems, including the perception module, planning module, communication module, control module, navigation module, human-drone interaction module, and payload module, highlighting their roles and contributions to UAV functionality, as demonstrated in Figure \ref{fig_functions}. The embodied configuration aspect focuses on the structural characteristics of UAV systems, covering the designs and applications of fixed-wing UAVs\cite{panagiotou2020aerodynamic}, multirotor UAVs \cite{villa2020survey,rashad2020fully}, unmanned helicopters \cite{alvarenga2015survey}, and hybrid UAVs \cite{saeed2018survey}. Furthermore, focusing on swarm intelligence for UAVs, this section summarizes the advancements in UAV swarm technologies, including communication strategies, formation control methods, and collaborative decision-making mechanisms.

\subsection{Functional Modules of UAVs}
\subsubsection{Perception Module}
The Perception Module serves as the UAV’s “eyes and ears,” collecting and interpreting data from a variety of onboard sensors to build a comprehensive understanding of the surrounding environment. These sensors include RGB cameras, event-based cameras, thermal cameras, 3D cameras, LiDAR, radar, and ultrasonic sensors \cite{du2024advancements}. By converting raw sensor data into actionable insights, such as detecting obstacles, identifying landmarks, and assessing terrain features. The Perception Module provides the situational awareness essential for safe and autonomous flight \cite{martinez2020review, zhang2023aerial}.

Beyond basic environmental monitoring, the Perception Module also supports collaborative tasks in multi-UAV operations, including the detection and tracking of other drones to facilitate coordinated swarm behavior. Advanced computer vision and machine learning techniques play a pivotal role in this process, enhancing the accuracy and robustness of object detection \cite{mittal2020deep,liu2020uav}, semantic segmentation \cite{ girisha2019semantic,liu2021light}, and motion estimation \cite{li2010vision,dobrokhodov2006vision}. Sensor fusion methods are often employed to combine complementary data sources, such as fusing LiDAR depth maps with high-resolution camera imagery, thereby mitigating the limitations of individual sensors while capitalizing on their unique strengths \cite{mascaro2018gomsf, wan2022uav}. This robust, multimodal perception framework enables UAVs to adapt to changing conditions (e.g., varying lighting, dynamic environments) and carry out complex missions with minimal human intervention.

\subsubsection{Navigation Module}
The Navigation Module is responsible for translating the planned trajectories from the Planning Module into precise flight paths by continuously estimating and adjusting the UAV’s position, orientation, and velocity \cite{rezwan2022artificial}. To achieve this, it relies on a variety of onboard sensors \cite{gyagenda2022review}, such as GPS \cite{balamurugan2016survey, mcenroe2022survey}, inertial measurement units \cite{neumann2015real, barbieri2019intercomparison}, visual odometry, and barometric sensors or magnetometers to gather real-time information about the UAV’s state \cite{couturier2021review, rovira2022review}. Sensor-fusion algorithms, including Kalman filters (e.g., Extended or Unscented Kalman Filters) and particle filters, are employed to integrate data from disparate sources, enhancing the reliability and accuracy of state estimation.

In GPS-denied or cluttered environments, the Navigation Module may employ simultaneous localization and mapping techniques or visual SLAM to provide robust localization and environment mapping \cite{atif2021uav, couturier2021review, lu2018survey, gupta2022simultaneous, kassas2024aircraft}. Such advanced solutions enable the UAV to maintain a high level of situational awareness even when traditional satellite-based positioning is unavailable or unreliable. By ensuring accurate state estimation and smooth trajectory tracking, the Navigation Module plays a critical role in maintaining overall flight stability and guaranteeing that the UAV adheres to the mission plan throughout its operational timeframe.

\subsubsection{Planning Module}
The Planning Module is pivotal in translating high-level mission objectives into concrete flight trajectories and actions, relying on input from the Perception Module to ensure safe navigation \cite{tisdale2009autonomous, goerzen2010survey}. Path-planning algorithms span a broad range of techniques aimed at computing feasible and often optimized routes around obstacles. These methods include heuristic algorithms such as the $A*$ algorithm \cite{hong2021quadrotor}, Evolutionary Algorithms \cite{chai2022multi}, SA \cite{xiao2021simulated, ait2022novel}, PSO \cite{phung2021safety, yu2022novel, he2021novel}, Pigeon-Inspired Optimization \cite{yang2023uav}, Artificial Bee Colony \cite{liu2021multi, han2022improved}, etc. Machine learning approaches, including Neural Networks \cite{pan2021deep, cui2021uav, heidari2023machine}, and Deep Reinforcement Learning \cite{he2021explainable, zhu2021uav}are also employed for more adaptive and data-driven planning. Additionally, sampling-based strategies like Rapidly-exploring Random Trees offer flexible frameworks for dealing with high-dimensional or dynamically changing environments \cite{guo2023hpo}. By leveraging one or a combination of these methods, UAVs are able to devise safe, collision-free trajectories that optimize key performance metrics, such as travel time, energy consumption, or overall mission efficiency. \cite{lin2017sampling, puente2021using, yang2023uav, puente2022review}. These techniques enable UAVs to operate autonomously within complex or uncertain environments by continuously adapting their planned path in real-time, particularly important when unforeseen changes occur in terrain, obstacle locations, or mission parameters.

In multi-UAV or swarm operations, the Planning Module also plays a key role in coordinating flight routes among individual drones, ensuring collision avoidance and maintaining cohesive group behaviors \cite{pan2021improved, zhao2021multi, li2024multi}. This collaborative planning capability not only enhances mission efficiency but also reduces the risk of inter-UAV interference. By dynamically updating trajectories and sharing relevant information, the Planning Module underpins robust, reliable operations that align with overall mission goals.

\subsubsection{Control Module}
The Control Module is responsible for generating low-level commands that regulate the UAV’s actuators—including motors, servos, and other control surfaces—to maintain stable and responsive flight. Acting as the “muscle” of the system, it continuously adjusts key parameters such as altitude, velocity, orientation, and attitude in response to real-time feedback from onboard sensors. By closing the control loop with reference inputs provided by the Navigation and Planning Modules, the Control Module ensures that the UAV adheres to desired flight trajectories and mission objectives \cite{fahlstrom2022introduction, harvey2022review}.

To manage potential disturbances (e.g., wind gusts, payload variations) and compensate for modeling uncertainties, a variety of classical and modern control strategies are employed. Traditional approaches, such as Proportional–Integral–Derivative control \cite{mahmoodabadi2020fuzzy, bello2022fixed}, offer simplicity and ease of implementation, while more advanced techniques like Model Predictive Control enable predictive action based on system dynamics and constraints. Adaptive control methods further enhance performance by adjusting control parameters in real time as the characteristics of the system evolve \cite{koksal2020backstepping, zuo2022unmanned}. Other robust strategies, such as sliding-mode control or nonlinear control can be used for particularly challenging operating conditions, providing resilience against sensor noise and sudden environmental changes \cite{fei2021fuzzy, gambhire2021review}.

In multi-rotor UAVs, for example, the Control Module finely tunes individual motor speeds to achieve the appropriate thrust and torque distributions for stable flight, whereas in fixed-wing platforms, it manipulates aerodynamic surfaces to maintain or alter flight paths \cite{jasim2020robust, basiri2022survey, boroujeni2024comprehensive}. This tight integration of sensor feedback, control algorithms, and actuator commands allows the UAV to respond quickly to deviations and external perturbations, ensuring smooth and reliable operations throughout the mission.

\subsubsection{Communication Module}

The Communication Module underpins all data exchanges between the UAV, ground control stations (GCS), and other external entities, such as satellites, edge devices, or cloud-based services, ensuring that critical telemetry, control, and payload information flows seamlessly. Typical communication methods range from short-range radio frequency systems and Wi-Fi links to more sophisticated, longer-range networks like 4G, 5G, or even satellite-based links, each selected to meet the specific mission requirements regarding bandwidth, latency, and range \cite{campion2018uav, sharma2020communication, hentati2020comprehensive, wu2021comprehensive}.

In UAV swarm operations, the Communication Module becomes particularly vital, it relays commands to and from ground control and facilitates inter-UAV coordination by sharing situational data (e.g., positions, sensor readings) in real time \cite{wu2021comprehensive, ullah2020cognition}. Robust communication protocols often augmented with encryption and authentication mechanisms guard against unauthorized access and malicious interference, while techniques like adaptive channel selection and multi-hop ad-hoc routing can mitigate signal degradation and ensure reliable connectivity in dynamic environments \cite{alladi2020applications}. By managing and prioritizing different data streams (telemetry, payload, command and control), the Communication Module serves as the backbone that keeps all subsystems in sync and supports the UAV’s overall operational objectives \cite{kumar2021sp2f}.

\subsubsection{Interaction Module}
The Interaction Module is designed to facilitate seamless communication and collaboration between the UAV and human operators or other agents in the operating environment \cite{messaoudi2023survey, yoo2022motion}. It encompasses user interfaces and interaction paradigms that may include voice commands, gesture recognition, augmented or virtual reality displays, or touchscreen-based data visualization systems \cite{li2021uav, sun2022human, zhang2020rfhui, deng2023vr,xiao2024macns}. Additional methods such as adaptive user interface design that tailors the displayed information to the operator’s skill level and workload, or haptic feedback mechanisms that provide tactile alerts for critical events can further enhance situational awareness and user experience \cite{jiao2020intuitive}. These interfaces enable ground personnel to issue high-level commands, review mission progress, and intervene when necessary, ensuring that operators maintain oversight and decision-making authority \cite{divband2021designing}.

In swarm or multi-UAV contexts, the Interaction Module becomes even more integral. It not only allows central decision makers to coordinate multiple drones but also enables human operators to receive aggregated situational data from across the swarm, potentially flagging anomalies or emergent behaviors in real time. These human-UAV interaction channels are particularly critical in collaborative missions (for example, search and rescue, environmental monitoring, or infrastructure inspection), where on-the-spot guidance or feedback may be required to adapt the UAVs’ behavior to evolving conditions \cite{zheng2020evolutionary, lim2021adaptive, chang2020coactive}. By providing robust mechanisms for manual overrides and real-time communication, the Interaction Module strikes a balance between autonomous operation and human-in-the-loop supervision, enhancing both mission effectiveness and operational safety \cite{lim2021adaptive, cauchard2021toward, ribeiro2021web}.

\subsubsection{Payload Module}
The Payload Module oversees the equipment or cargo the UAV carries to accomplish its mission objectives. Depending on the task, these payloads may range from cameras for surveillance, to delivery packages, to advanced sensors for environmental monitoring, to specialized hardware for tasks such as search and rescue \cite{mohiuddin2023dual}. Consequently, the Payload Module must address a variety of operational needs, including power supply, secure data transmission, mechanical support, and proper stabilization to ensure reliable performance under diverse conditions \cite{fahlstrom2022introduction, gonzalez2017unmanned}.

In practice, this module often integrates features such as vibration damping, thermal management, and secure mounting solutions to protect delicate components and maintain optimal functionality during flight. Moreover, in some UAV designs, the Payload Module is designed to be interchangeable. This modular approach, which typically employs standardized mounting and connectivity interfaces, enables rapid swapping of payloads and streamlines the process of configuring a UAV for different mission profiles. As a result, operators can expand the drone’s capabilities without requiring an entirely new platform, thereby enhancing flexibility and reducing both deployment time and cost \cite{hadi2014autonomous, kusznir2020sliding, lee2020antisway, mohammadi2020control} .

Overall, the Payload Module plays a crucial role in bridging the UAV’s core flight systems with the mission-specific tools essential for achieving operational objectives. By accommodating a wide range of payloads and ensuring they are powered, protected, and efficiently connected, the Payload Module significantly extends the UAV’s applicability across various industries and mission types.

\subsection{Embodied Configurations of UAVs}

UAVs can be categorized into several types based on their geometric configurations. These include fixed-wing UAVs, multirotor UAVs, unmanned helicopters, and other types. Below, we introduce these categories and summarize their characteristics.

\subsubsection{Fixed-Wing UAVs}

Fixed-wing UAVs feature a predetermined wing shape that generates lift as air flows over the wings, enabling forward motion \cite{gonzalez2017unmanned}. These UAVs are known for their high speed, long endurance, and stable flight, making them ideal for long-duration missions. However, they require advanced piloting skills and cannot perform hovering \cite{lee2021survey}. Fixed-wing UAVs are commonly used for monitoring fields, forests, highways, and railways \cite{gonzalez2017unmanned}.

\subsubsection{Multirotor UAVs}

Multirotor UAVs are one of the most prevalent types in daily life, typically equipped with multiple rotors (commonly four, six, or more) to generate lift through rotor rotation. Their advantages include low cost, ease of operation, and the ability for vertical take-off and landing (VTOL) and hovering, making them suitable for precision tasks. However, they have limited endurance and relatively low speed. Multirotor UAVs are often used for tasks such as photography, agricultural monitoring, and spraying.

\subsubsection{Unmanned Helicopters}

Unmanned helicopters are equipped with one or two powered rotors to provide lift and enable attitude control. This design allows for vertical take-off, hovering, and high maneuverability. Compared to multirotor UAVs, unmanned helicopters have superior payload capacity, enabling them to carry heavier equipment or sensors. Their strengths include long endurance and excellent wind resistance, making them stable even in strong winds. The main limitation is their relatively low speed. Unmanned helicopters find widespread applications in areas such as traffic surveillance, resource exploration, forest fire prevention, and military reconnaissance.

\subsubsection{Hybrid UAV}

Hybrid UAVs combine the strengths of both fixed-wing and multirotor UAVs, offering a versatile design that allows for VTOL while also achieving the long endurance and high speed typical of fixed-wing UAVs. These UAVs typically feature a combination of rotors for lift during vertical flight and wings for sustained forward motion. The main advantage of hybrid UAVs is their flexibility, enabling them to perform a wide range of missions, including those requiring both hovering and long-duration flight. However, the complexity of their design and mechanisms results in higher costs and more demanding maintenance.

\subsubsection{Flapping-Wing UAV}

Flapping-wing UAVs are bio-inspired unmanned aerial vehicles that mimic the flight mechanisms of birds or insects. These UAVs rely on unsteady aerodynamic effects generated by wing flapping to achieve flight. They offer quieter operation, higher efficiency, and increased maneuverability compared to conventional UAVs. Their compact size is a notable advantage, but they generally have a lower payload capacity. Additionally, the design and control systems of flapping-wing UAVs are more complex due to the dynamic nature of their flight mechanics.

\subsubsection{Unmanned Airship}

Unmanned airships are a type of aerial vehicle that utilizes lightweight gases for buoyancy and employs propulsion and external structural elements for movement and directional control. These airships are highly cost-effective and produce low flight noise. However, their agility is limited, and they operate at relatively low speeds. Due to their large size, unmanned airships are highly susceptible to wind influences, which can affect their stability and operational reliability.

\begin{table*}[htbp]
    \centering
    \caption{Typical configurations of UAV}
     \setlength{\tabcolsep}{5pt} 
    \renewcommand{\arraystretch}{1.5} 
    \scriptsize 
    \newcolumntype{C}[1]{>{\centering\arraybackslash}m{#1}}

    \begin{tabular}{@{}m{3cm}m{5cm}m{4.5cm}m{4.5cm}@{}}
        \toprule
        \textbf{Category}         & \textbf{Characteristics}                                                                                     & \textbf{Advantages}                                                                 & \textbf{Disadvantages}                                     \\ \midrule
        Fixed-wing UAV     & Fixed wings generate lift with forward motion.                                    & High speed, long endurance, stable flight.                                     &  Cannot hover, high demands for takeoff/landing areas.            \\

        Multirotor UAV      & Multiple rotors provide lift and control.                                            & Low cost, easy operation, capable of VTOL and  hovering.                                 & Limited flight time, low speed, small payload capacity.                        \\

        Unmanned Helicopter & Single or dual rotors allow vertical take-off and hovering.                                & High payload capacity, good wind resistance, long endurance, VTOL.                 & Complex structure, higher maintenance cost, slower than fixed-wing UAVs.                                           \\

        Hybrid UAV & Combines fixed-wing and multirotor capabilities. & Flexible missions, long endurance, VTOL. & Complex mechanisms, higher cost.\\

        Flapping-wing UAV  & Uses clap-and-fling mechanism for flight. & Low noise, high propulsion efficiency, high maneuverability. & Complex analysis and control, limited payload capacity. \\

        Unmanned airship & Aerostat aircraft with gasbag for lift.  & Low cost, low noise. & low speed, low maneuverability, highly affected by wind. \\
         \bottomrule
    \end{tabular}
    \label{tab:uav_types}
\end{table*}

\subsection{UAV Swarm}
UAV swarms involve multiple UAVs working collaboratively to achieve a shared objective, offering advantages in redundancy, scalability, and efficiency compared to individual UAV operations\cite{zhou2020uav}. The swarm approach relies on decentralized decision-making, allowing UAVs to adjust their behaviors in response to the actions of their peers and environmental changes. Swarm algorithms often draw inspiration from biological systems\cite{chakraborty2017swarm,jiao2024nature}, such as flocks of birds or colonies of ants, utilizing techniques like consensus algorithms\cite{lamport2001paxos}, PSO\cite{kennedy1995particle}, or behavior-based coordination\cite{jones2018behavior}. Effective swarm operation requires seamless communication, robust control mechanisms, and cooperative planning to manage the complexities of distributed systems\cite{ma2022survey}. This concept is particularly useful in applications like large-area surveillance, precision agriculture, and search and rescue, where multiple UAVs can cover a greater area more efficiently than a single vehicle.

In this section, we will discuss key components essential for effective UAV swarm operation, including task allocation, communication architecture, path planning, and formation control.

\subsubsection{Task allocation in UAV swarm}

Task allocation in UAV swarms involves efficiently distributing tasks among multiple UAVs to optimize mission performance \cite{schwarzrock2018solving}. This allocation problem is NP-hard, with complexity growing exponentially with swarm size and task count \cite{zhang2021uav}. It is typically modeled as the Traveling Salesman Problem (TSP) \cite{kudo2023tsp}, Vehicle Routing Problem (VRP) \cite{sarkar2018scalable}, Mixed-Integer Linear Programming (MILP) \cite{darrah2005multiple}, or Cooperative Multi-task Allocation Problem (CMTAP) \cite{ye2020cooperative}. Common approaches include heuristic algorithms, AI-based methods, mathematical programming, and market-based mechanisms.

Heuristic algorithms, such as Genetic Algorithms (GAs), Particle Swarm Optimization (PSO), and Simulated Annealing (SA), efficiently search for feasible solutions without easily falling into local optima. Han \emph{et al.} \cite{han2021modified} developed a Fuzzy Elite Strategy Genetic Algorithm (FESGA), while Yan \emph{et al.} \cite{yan2024cooperative} proposed an enhanced GA for integrated task allocation and path planning.

PSO algorithms effectively balance exploration and exploitation, offering simplicity and speed. Jiang \emph{et al.} \cite{jiang2017method} introduced an improved PSO for multi-constraint task allocation, and Gao \emph{et al.} \cite{gao2018multi} used a Multi-Objective PSO (MOPSO) for multi-UAV allocation.

Mathematical programming approaches provide precise optimal solutions but become computationally expensive for larger problems. For instance, Choi \emph{et al.} \cite{choi2010task} formulated the UAV task allocation as a MILP model.

AI-based methods, such as reinforcement learning and neural networks, dynamically adapt to changing environments. Yang \emph{et al.} \cite{yang2019application} presented a reinforcement learning-based task scheduling algorithm, and Yin \emph{et al.} \cite{yin2022task} applied deep transfer reinforcement learning to multi-UAV task allocation.

Market-based methods, including auction algorithms and the Contract Net Protocol (CNP) \cite{peng2021review}, leverage incentives for efficient distributed allocation \cite{skaltsis2023review}. Examples include the auction-based approach proposed by Qiao \emph{et al.} \cite{cheng2016auction}, the hierarchical auction algorithm developed by Duan \emph{et al.} \cite{duan2019novel}, which addresses heterogeneous task allocation and obstacle avoidance, the hybrid contract net protocol (CNP) method presented by Zhang \emph{et al.} \cite{zhang2022dynamic}, and the two-stage distributed task allocation algorithm introduced by Wang \emph{et al.} \cite{wang2023two}.

\subsubsection{Communication architecture in UAV swarm}

For UAV swarms, communication is essential for coordination, enabling collaborative work and maintaining stability during operations. Communication can be achieved through two main approaches: infrastructure-based architectures \cite{campion2018review} and Flying Ad-hoc Network (FANET) architectures \cite{bekmezci2013flying}. Each method offers unique advantages and challenges, which will be discussed below.

Infrastructure-based Architectures:
This architecture depends on GCS \cite{campion2018uav} to manage the swarm. The GCS collects telemetry data from UAVs and transmits commands, either in real-time or through pre-programmed instructions. Its key advantages include centralized computation and real-time optimization, eliminating the need for inter-drone communication networks \cite{campion2018review}. However, this approach has notable limitations: the entire system is vulnerable to single-point failures in the GCS, UAVs must remain within the GCS communication range, and the architecture lacks the flexibility of distributed decision-making \cite{campion2018review}.

FANET Architecture:
FANETs consist of UAVs communicating directly with one another without needing a central access point. This decentralized network enables UAVs to coordinate tasks autonomously, with at least one UAV maintaining a link to a ground base or satellite. FANETs benefit from flexibility, scalability, and reduced dependency on infrastructure. However, they require robust communication protocols and may face challenges in managing dynamic topologies and ensuring reliability \cite{bekmezci2013flying}.

\subsubsection{Path planning in UAV swarm}

UAV swarm path planning refers to selecting an optimal path for the UAV swarm from the starting position to all target positions, while ensuring the predefined distance between UAVs to avoid collisions \cite{javed2024state}. The optimal path generally refers to the shortest path length, shortest travel time, least energy consumption, and other event-specific constraints \cite{javed2024state}. The criteria for the optimal path need to be determined based on the actual problem. UAV path planning algorithms can generally be divided into three major categories: intelligent optimization algorithms, mathematical programming methods, and AI-based approaches. Below, we briefly introduce these three methods.

In nature, various group behaviors, such as flocks of birds, schools of fish, and ant colonies, follow specific rules that enable efficient food searching or migration. These behaviors can be abstracted into mathematical models for information transfer, path planning, and coordinated control, which are applicable to UAV swarm scheduling. Common intelligent optimization algorithms for UAV swarms include Ant Colony Optimization (ACO), GAs, SA, and PSO. For instance, ACO mimics the foraging behavior of ants, where ants probabilistically select paths based on pheromone concentration, ultimately finding optimal or near-optimal solutions. Researchers such as Turker \emph{et al.} \cite{turker2016gpu} have applied SA to UAV swarm path planning, while Wei \emph{et al.} \cite{wei2009path} used ACO for the same purpose.

Beyond heuristic algorithms inspired by nature, mathematical models like MILP and Nonlinear Programming can be directly applied to UAV swarm scheduling for precise solutions. For example, Ragi \emph{et al.} \cite{ragi2017mixed} used Mixed-Integer Nonlinear Programming (MINLP) to address UAV path planning. While these methods are effective for small-scale problems, their computational complexity increases exponentially as the problem size grows.

With the rise of machine learning, AI-based algorithms have also been applied to UAV swarm scheduling and optimization. Kool \emph{et al.} \cite{kool2018attention} used deep learning for vehicle routing, and similar approaches have been adopted in UAV swarm path planning. Xia \emph{et al.} \cite{xia2018multi} applied neural networks to UAV path planning, Sanna \emph{et al.} \cite{sanna2021neural} extended this to multi-UAV planning, and Puente-Castro \emph{et al.} \cite{puente2021using} applied reinforcement learning. By training on extensive datasets, neural networks can learn to model the environment, including obstacles and their dynamic changes, thereby improving the accuracy of path planning.

\subsubsection{UAV Swarm Formation Control Algorithm}

The UAV swarm relies on effective formation control algorithms that enable it to autonomously form and maintain a formation to perform tasks, and switch or rebuild the formation based on specific tasks \cite{ouyang2023formation}. The primary approaches to formation control are centralized, decentralized, and distributed control algorithms \cite{bu2024advancement}.

Centralized Control: Centralized control involves a central unit that oversees task allocation and resource management, with individual UAVs primarily responsible for data input, output, and storage \cite{ouyang2023formation}. This approach simplifies decision-making, ensures coordinated actions, and is relatively straightforward to implement. However, it is susceptible to high communication overhead and single points of failure; if the central unit fails, the entire system may collapse. Common methods in centralized control include virtual structure \cite{askari2015uav} and leader-follower approaches \cite{lewis1997high, desai1998controlling}.

Decentralized Control: In decentralized control, each UAV makes decisions based on local sensors and controllers, without requiring explicit communication with other UAVs \cite{huang2022decentralized}. UAVs adjust their movements to maintain formation based on local conditions and predefined rules. The primary advantages of this approach include flexibility and ease of adapting formations. However, the lack of access to global information results in poor control performance, requiring continuous iteration \cite{sun2022observation}.

Distributed Control: Distributed control involves extensive communication between UAVs, enabling them to coordinate and maintain formation through shared information. UAVs work collaboratively to make optimal decisions based on both local data and pre-established rules. Compared to decentralized control, distributed control benefits from more robust collaboration and improved flexibility. However, it requires higher communication demands and more complex algorithms to manage coordination, which increases both the computational burden and the risk of communication failures. Typical methods include behavior method \cite{duan2021homing} and consistency method \cite{tao2023multi}.

\section{Preliminaries on FMs}
\label{sec:Preliminaries_on_Foundation_models}

This section provides an overview of FMs, including LLMs, Vision Foundation Models (VFMs), and Vision Language Models (VLMs). It highlights their core characteristics and technical advantages, with the aim of offering foundational insights and guidance for the deep integration of these models with UAV systems.

\renewcommand{\arraystretch}{1.2} 
\setlength{\tabcolsep}{4pt} 
\begin{table*}[htbp]
    \centering
    \caption{Summarization of LLMs, VLMs, and VFMs.}
    \label{tab:llms}
    \setlength{\tabcolsep}{5pt} 
    \renewcommand{\arraystretch}{1} 
    \scriptsize 
    \newcolumntype{C}[1]{>{\centering\arraybackslash}m{#1}}

    \begin{tabular}{@{}C{3cm}C{2cm}C{7cm}C{4cm}@{}}
        \toprule
        \textbf{Category} & \textbf{Subcategory} & \textbf{Model Name} & \textbf{Institution / Author} \\
        \midrule
        \multirow{11}{*}{LLMs} 
        & \multirow{11}{*}{General} 
        & GPT-3\cite{brown2020language}, GPT-3.5\cite{ouyang2022training}, GPT-4\cite{achiam2023gpt} 
        & \href{https://openai.com}{OpenAI} \\
        &  & Claude 2, Claude 3\cite{claude_model_card, constitutional_ai, mapping_mind_llm} 
        & \href{https://www.anthropic.com}{Anthropic} \\
        &  & Mistral series\cite{jiang2023mistral, jiang2024mixtral} 
        & \href{https://www.mistral.ai}{Mistral AI} \\
        &  & PaLM series\cite{chowdhery2023palm, driess2023palm}, Gemini series\cite{team2023gemini, reid2024gemini} 
        & \href{https://research.google}{Google Research} \\
        &  & LLaMA\cite{touvron2023llama}, LLaMA2\cite{touvron2023llama2}, LLaMA3\cite{dubey2024llama} 
        & \href{https://www.llama.com}{Meta AI} \\
        &  & Vicuna\cite{chiang2023vicuna} 
        & \href{https://vicuna.lmsys.org}{Vicuna Team} \\
        &  & Qwen series\cite{bai2023qwen, yang2024qwen2} 
        & \href{https://github.com/QwenLM}{Qwen Team, Alibaba Group} \\
        &  & InternLM\cite{team2023internlm, cai2024internlm2} 
        & \href{https://github.com/InternLM/InternLM}{Shanghai AI Laboratory} \\
        &  & BuboGPT\cite{zhao2023bubogpt} 
        & \href{https://github.com/magic-research/bubogpt}{Bytedance} \\
        &  & ChatGLM\cite{du2021glm, zeng2022glm, glm2024chatglm} 
        & \href{https://github.com/THUDM}{THUKEG\&THUDM} \\
        &  & DeepSeek series\cite{bi2024deepseek, liu2024deepseek, guo2024deepseek, guo2025deepseek} 
        & \href{https://github.com/deepseek-ai}{DeepSeek} \\
        \midrule
        \multirow{18}{*}{VLMs}
        & \multirow{10}{*}{General}
        & GPT-4V\cite{openai2024gpt4v}, GPT-4o, GPT-4o mini, GPT o1-preview
        & \href{https://openai.com}{OpenAI} \\
        &  & Claude 3 Opus, Claude 3.5 Sonnet\cite{claude3modelcard2024}
        & \href{https://www.anthropic.com}{Anthropic} \\
        &  & Step-2
        & \href{https://www.stepfun.com/}{Jieyue Xingchen} \\
        &  & LLaVA\cite{liu2024visual}, LLaVA-1.5\cite{liu2024improved}, LLaVA-NeXT\cite{liu2024llavanext}
        & \href{https://github.com/haotian-liu/LLaVA}{Liu \emph{et al}.} \\
        &  & MoE-LLaVA\cite{lin2024moe}
        & \href{https://github.com/PKU-YuanGroup/MoE-LLaVA}{Lin \emph{et al}.} \\
        &  & LLaVA-CoT\cite{xu2024llava}
        & \href{https://github.com/PKU-YuanGroup/LLaVA-CoT}{Xu \emph{et al}.} \\
        &  & Flamingo\cite{alayrac2022flamingo}
        & {Alayrac \emph{et al}.} \\
        &  & BLIP\cite{li2022blip}
        & \href{https://github.com/salesforce/BLIP}{Li \emph{et al}.} \\
        &  & BLIP-2\cite{li2023blip2}
        & \href{https://github.com/salesforce/LAVIS/tree/main/projects/blip2}{Li \emph{et al}.} \\
        &  & InstructBLIP\cite{dai2023instructblip}
        & \href{https://github.com/salesforce/LAVIS/tree/main/projects/instructblip}{Dai \emph{et al}.} \\
        \cmidrule(lr){2-4}
        & \multirow{4}{*}{Video Understanding}
        & LLaMA-VID\cite{li2025llama}
        & \href{https://github.com/dvlab-research/LLaMA-VID}{Li \emph{et al}.} \\
        &  & IG-VLM\cite{kim2024image}
        & \href{https://github.com/imagegridworth/IG-VLM}{Kim \emph{et al}.} \\
        &  & Video-ChatGPT\cite{maaz2023video}
        & \href{https://github.com/mbzuai-oryx/Video-ChatGPT}{Maaz \emph{et al}.} \\
        &  & VideoTree\cite{wang2024videotree}
        & \href{https://github.com/Ziyang412/VideoTree}{Wang \emph{et al}.} \\
        \cmidrule(lr){2-4}
        & \multirow{4}{*}{Visual Reasoning}
        & X-VLM\cite{zeng2021multi}
        & \href{https://github.com/zengyan-97/X-VLM}{Zeng \emph{et al}.} \\
        &  & Chameleon\cite{lu2024chameleon}
        & \href{https://chameleon-llm.github.io/}{Lu \emph{et al}.} \\
        &  & HYDRA\cite{ke2024hydra}
        & \href{https://hydra-vl4ai.github.io/}{Ke \emph{et al}.} \\
        &  & VISPROG\cite{gupta2023visual}
        & \href{https://prior.allenai.org/projects/visprog}{PRIOR @ Allen Institute for AI} \\
        \midrule
        \multirow{31}{*}{VFMs}
        & \multirow{4}{*}{General}
        & CLIP\cite{radford2021learning}
        & \href{https://github.com/OpenAI/CLIP}{OpenAI} \\
        &  & FILIP\cite{yao2021filip}
        & Yao \emph{et al}. \\
        &  & RegionCLIP\cite{zhong2022regionclip}
        & \href{https://github.com/microsoft/RegionCLIP}{Microsoft Research} \\
        &  & EVA-CLIP\cite{sun2023eva}
        & \href{https://github.com/baaivision/EVA/tree/master/EVA-CLIP}{Sun \emph{et al}.} \\
        \cmidrule(lr){2-4}
        & \multirow{7}{*}{Object Detection}
        & GLIP\cite{li2022grounded}
        & \href{https://github.com/microsoft/GLIP}{Microsoft Research} \\
        &  & DINO\cite{zhang2022dino}
        & \href{https://github.com/IDEA-Research/DINO}{Zhang \emph{et al}.} \\
        &  & Grounding-DINO\cite{liu2023grounding}
        & \href{https://github.com/IDEA-Research/GroundingDINO}{Liu \emph{et al}.} \\
        &  & DINOv2\cite{oquab2023dinov2}
        & \href{https://github.com/facebookresearch/dinov2}{Meta AI Research, FAIR} \\
        &  & AM-RADIO\cite{ranzinger2024radio}
        & \href{https://github.com/NVlabs/RADIO}{NVIDIA} \\
        &  & DINO-WM\cite{zhou2024dino}
        & \href{https://dino-wm.github.io/}{Zhou \emph{et al}.} \\
        &  & YOLO-World\cite{cheng2024yolo}
        & \href{https://github.com/AILab-CVC/YOLO-World}{Cheng \emph{et al}.} \\
        \cmidrule(lr){2-4}
        & \multirow{15}{*}{Image Segmentation}
        & CLIPSeg\cite{luddecke2022image}
        & \href{https://github.com/timojl/clipseg}{Lüdecke and Ecker} \\
        &  & SAM\cite{kirillov2023segment}
        & \href{https://segment-anything.com}{Meta AI Research, FAIR} \\
        &  & Embodied-SAM\cite{xu2024embodiedsam}
        & \href{https://github.com/xuxw98/ESAM}{Xu \emph{et al}.} \\
        &  & Point-SAM\cite{zhou2024point}
        & \href{https://github.com/zyc00/Point-SAM}{Zhou \emph{et al}.} \\
        &  & Open-Vocabulary SAM\cite{yuan2025open}
        & \href{https://www.mmlab-ntu.com/project/ovsam/}{Yuan \emph{et al}.} \\
        &  & TAP\cite{pan2025tokenize}
        & \href{https://github.com/baaivision/tokenize-anything}{Pan \emph{et al}.} \\
        &  & EfficientSAM\cite{xiong2024efficientsam}
        & \href{https://yformer.github.io/efficient-sam/}{Xiong \emph{et al}.} \\
        &  & MobileSAM\cite{zhang2023faster}
        & \href{https://github.com/ChaoningZhang/MobileSAM}{Zhang \emph{et al}.} \\
        &  & SAM 2\cite{ravi2024sam}
        & \href{https://ai.meta.com/sam2/}{Meta AI Research, FAIR} \\
        &  & SAMURAI\cite{yang2024samurai}
        & \href{https://github.com/yangchris11/samurai}{University of Washington} \\
        &  & SegGPT\cite{wang2023seggpt}
        & \href{https://github.com/baaivision/Painter/tree/main/SegGPT}{Wang \emph{et al}.} \\
        &  & Osprey\cite{yuan2024osprey}
        & \href{https://github.com/CircleRadon/Osprey}{Yuan \emph{et al}.} \\
        &  & SEEM\cite{zou2024segment}
        & \href{https://github.com/UX-Decoder/Segment-Everything-Everywhere-All-At-Once}{Zou \emph{et al}.} \\
        &  & Seal\cite{liu2024segment}
        & \href{https://github.com/youquanl/Segment-Any-Point-Cloud}{Liu \emph{et al}.} \\
        &  & LISA\cite{lai2024lisa}
        & \href{https://github.com/dvlab-research/LISA}{Lai \emph{et al}.} \\
        \cmidrule(lr){2-4}
        & \multirow{5}{*}{Depth Estimation}
        & ZoeDepth\cite{bhat2023zoedepth}
        & \href{https://github.com/isl-org/ZoeDepth}{Bhat \emph{et al}.} \\
        &  & ScaleDepth\cite{zhu2024scaledepth}
        & \href{https://ruijiezhu94.github.io/ScaleDepth/}{Zhu \emph{et al}.} \\
        &  & Depth Anything\cite{yang2024depth1}
        & \href{https://depth-anything.github.io}{Yang \emph{et al}.} \\
        &  & Depth Anything V2\cite{yang2024depth2}
        & \href{https://depth-anything-v2.github.io/}{Yang \emph{et al}.} \\
        &  & Depth Pro\cite{bochkovskii2024depth}
        & \href{https://github.com/apple/ml-depth-pro}{Apple} \\
        \bottomrule
    \end{tabular}
\end{table*}

\subsection{LLMs}

In recent years, LLMs have seen rapid advancements, with increasingly larger models being trained on diverse, large-scale corpora \cite{vaswani2017attention}. These models have consistently set new performance benchmarks in various NLP tasks and have been widely adopted in both academic research and industrial applications \cite{minaee2024large, zhao2023survey, chang2024survey, naveed2023comprehensive}. This section provides an overview of the core capabilities of LLMs, including their generalization and reasoning abilities, followed by an introduction to typical LLMs from leading research organizations. 

\subsubsection{Core Capabilities of LLMs}

\textbf{Generalization Capability:} Benefiting from training on large-scale corpora and the substantial size of the models, LLMs exhibit strong transfer capabilities, including zero-shot and few-shot learning. These capabilities enable LLMs to generalize effectively to new tasks, either without task-specific examples or with limited guidance, making them versatile tools for a wide range of applications. In zero-shot learning, without additional task-specific training, LLMs can solve relevant problems solely through natural language instructions. In few-shot learning, the model can quickly adapt to new tasks by leveraging several examples from the support set along with the corresponding task instructions \cite{li2023one}. 

The design of natural language instructions or prompts is crucial in enhancing generalization capability. Prompts not only provide natural language descriptions of tasks but also guide the model to perform tasks accurately based on input examples \cite{radford2019language, brown2020language, achiam2023gpt}. Furthermore, LLMs exhibit in-context learning, which allows them to learn and adapt to new tasks directly from the context provided within the prompt, such as task instructions and examples, without requiring any explicit retraining or model updates \cite{brown2020language, liu2021makes, dong2022survey}.

\textbf{Complex Problem-Solving Capability:} LLMs demonstrate a remarkable ability to solve complex problems by generating intermediate reasoning steps or structured logical pathways, facilitating a systematic and step-by-step approach to addressing challenges. This capability is exemplified by the Chain of Thought (CoT) framework, where intricate problems are decomposed into a series of manageable sub-tasks, each solved sequentially using examples of step-by-step reasoning \cite{kojima2022large, zhang2022automatic, wei2022chain, feng2024towards}. Besides, LLMs also demonstrate advanced capabilities in task planning and the orchestration of tools, enabling them to invoke appropriate resources to address specific sub-task requirements and efficiently integrate workflows to achieve comprehensive solutions \cite{shen2024hugginggpt, khot2022decomposed, huang2024understanding}.

\subsubsection{Typical LLMs}

Several notable milestones have marked the development of LLMs. OpenAI's GPT series, spanning GPT-3, GPT-3.5, and GPT-4, has set benchmarks in language understanding, generation, and reasoning tasks by leveraging extensive parameters and optimized architectures \cite{brown2020language, ouyang2022training, achiam2023gpt}. Anthropic's Claude models, including Claude 2 and Claude 3, prioritize safety and controllability through reinforcement learning, excelling in multi-task generalization and robustness \cite{claude_model_card, constitutional_ai, mapping_mind_llm}. The Mistral series employs sparse activation techniques to balance efficiency with performance, emphasizing low-latency inference \cite{jiang2023mistral, jiang2024mixtral}.

Google’s PaLM series \cite{chowdhery2023palm, driess2023palm} stands out for its multimodal capabilities and large-scale parameterization, while the subsequent Gemini series extends these features to improve generalizability and multilingual support \cite{team2023gemini, reid2024gemini}. In the open-source ecosystem, Meta's Llama models, including Llama 2 and Llama 3, excel in multilingual tasks and complex problem-solving. Derivative models like Vicuna enhance conversational abilities and task adaptability through fine-tuning on conversational datasets and techniques like Low-Rank Adaptation (LoRA) \cite{touvron2023llama, touvron2023llama2, dubey2024llama, chiang2023vicuna}. Similarly, the Qwen series, pre-trained on multilingual datasets and instruction-tuned, demonstrates adaptability across diverse tasks \cite{bai2023qwen}.

Several other models have achieved significant progress in specialized domains. InternLM \cite{cai2024internlm2}, BuboGPT \cite{zhao2023bubogpt}, ChatGLM \cite{du2021glm, zeng2022glm, glm2024chatglm}, and DeepSeek \cite{bi2024deepseek, liu2024deepseek, guo2024deepseek} focus on domain-specific tasks such as knowledge-based Q\&A, conversational generation, and information retrieval, enabled by task-specific fine-tuning and targeted extensions. Notably, LiveBench \cite{white2024livebench} has emerged as a comprehensive benchmarking platform, addressing limitations of previous evaluation standards. It systematically assesses LLMs' real-world capabilities across multi-task scenarios, offering valuable insights for model development and application.

\subsection{VLMs}
VLMs are multimodal models that extend the capabilities of LLMs by integrating visual and textual information\cite{ma2024llms}. These models are designed to tackle a range of tasks that require both vision and language understanding, such as visual question answering (VQA) and image captioning \cite{du2022survey, long2022vision, zhou2022learning, yin2024survey, zhang2024vision}. This section introduces several typical VLM models highlighting their technical features and application scenarios.

OpenAI's GPT-4V \cite{openai2024gpt4v} is a prominent representative in VLMs, demonstrating powerful visual perception capabilities \cite{yang2023dawn}. The upgraded GPT-4o introduces more advanced optimization algorithms, allowing it to accept arbitrary combinations of text, audio, and image inputs while delivering rapid responses. The lightweight version, GPT-4o mini, is designed for mobile devices and edge computing scenarios, balancing efficient performance with deployability by reducing computational resource consumption \cite{islam2024gpt}. GPT o1-preview excels in reasoning, particularly in programming and solving complex problems \cite{latif2024systematic}. Anthropic's Claude 3 Opus exhibits robust multi-task generalization and controllability, while Claude 3.5 Sonnet enhances practical value by optimizing reasoning speed and cost efficiency \cite{claude3modelcard2024}. The Step-2 model employs an innovative Mixture of Experts (MoE) architecture, supporting efficient training at a trillion-parameter scale and significantly improving the handling of complex tasks and model scalability.

Liu \emph{et al}. \cite{liu2024visual} proposed LLaVA, a representative VLM. This model leverages GPT-4 to generate instruction-following datasets and integrates CLIP visual encoder ViT-L/14 \cite{radford2021learning} with Vicuna \cite{vicuna2023}, fine-tuning end-to-end instruction to enhance its performance in multimodal tasks. Its latest version, LLaVA-NeXT \cite{liu2024llavanext}, builds upon LLaVA-1.5 \cite{liu2024improved} with significant improvements, notably enhancing the ability to capture visual details and excelling in complex visual and logical reasoning tasks. MoE-LLaVA replaces the language model in LLaVA with an MoE architecture, substantially improving inference efficiency and resource utilization in large-scale multi-task scenarios \cite{lin2024moe}. LLaVA-CoT enhances accuracy in reasoning-intensive tasks through structured reasoning annotations of large-scale visual question-answering samples combined with beam search methods \cite{xu2024llava}. Another important class of architectures includes the Flamingo \cite{alayrac2022flamingo} and BLIP series \cite{li2022blip, li2023blip2}, which enable LLMs to generate corresponding textual outputs from multimodal inputs by combining pre-trained visual feature encoders with pre-trained LLMs. Flamingo introduces the Perceiver Resampler and Gated Cross-Attention mechanisms, effectively integrating visual, multimodal information with the language model, thereby significantly enhancing performance in multimodal tasks. BLIP-2 \cite{li2023blip2} adopts a pretraining strategy combining stage-wise frozen image encoders with LLMs and introduces a Query Transformer (Q-Former) to effectively address alignment issues between the visual and language modalities. InstructBLIP \cite{dai2023instructblip} incorporates large-scale task instruction fine-tuning mechanisms, further improving the model's adaptability to multimodal tasks.

Additionally, VLMs have demonstrated broad application potential across various tasks and scenarios. In video understanding, representative models such as LLaMA-VID \cite{li2025llama}, IG-VLM \cite{kim2024image}, Video-ChatGPT \cite{maaz2023video}, and VideoTree \cite{wang2024videotree} exhibit outstanding performance in video content analysis and multimodal tasks. In visual reasoning, models like X-VLM \cite{zeng2021multi}, Chameleon \cite{lu2024chameleon}, HYDRA \cite{ke2024hydra}, and VISPROG \cite{gupta2023visual} enhance the accuracy and adaptability of complex visual reasoning tasks through innovative architectural designs and reasoning mechanisms.

\subsection{VFMs}

In recent years, the concept of VFMs has emerged as a core technology in computer vision. The primary goal of VFMs is to extract diverse and highly expressive image features, making them directly applicable to various downstream tasks. These models are typically characterized by large-scale parameters, remarkable generalization capabilities, and outstanding cross-task transfer performance, albeit with relatively high training costs \cite{ranzinger2024radio}. CLIP is a pioneering representative in the field of VFMs. By employing weakly supervised training on large-scale image-text pairs, it efficiently aligns visual and textual embeddings, laying a solid foundation for multimodal learning \cite{radford2021learning}. Subsequent works have further improved the training efficiency and performance of CLIP, including models such as FILIP \cite{yao2021filip}, RegionCLIP \cite{zhong2022regionclip}, and EVA-CLIP \cite{sun2023eva}.

\begin{figure}[ht]
 \centering
 \includegraphics[scale=0.6]{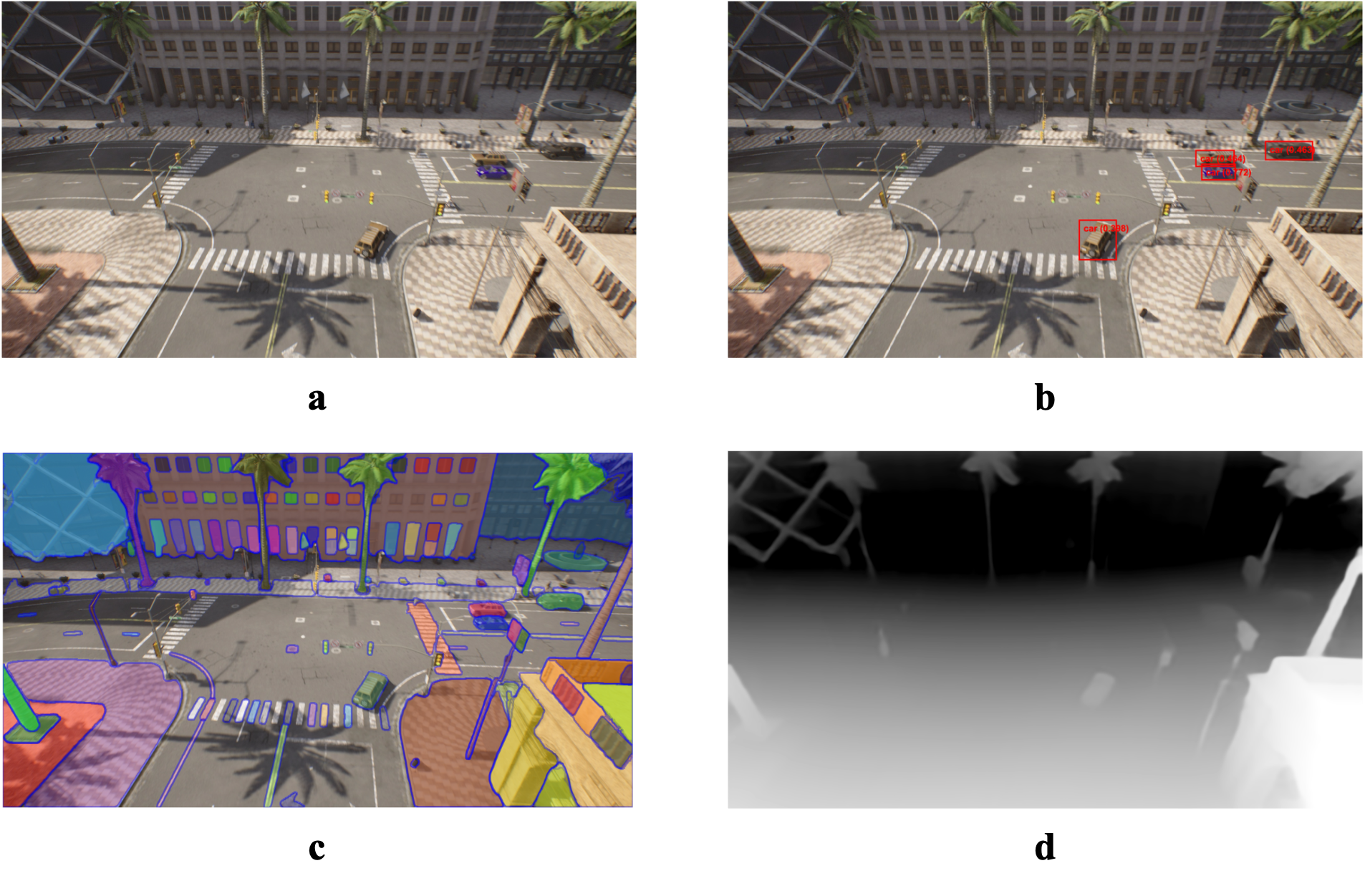}
 \caption{Demonstration of VFM models in various vision tasks. (a) The original image from the SynDrone dataset \cite{rizzoli2023syndrone}; (b) object detection result using Grounding DINO \cite{liu2023grounding} with the natural language prompt “car” as the detection target; (c) semantic segmentation of the entire image using the SAM model \cite{kirillov2023segment}; (d) Depth image generated for the entire image using the ZoeDepth model \cite{bhat2023zoedepth}.}
 \label{fig_1}
\end{figure}

VFMs have demonstrated exceptional adaptability, achieving remarkable results in various computer vision tasks, including zero-shot object detection, image segmentation, and depth estimation. As shown in Figure \ref{fig_1}, we selected a sample image from the Town10HD scene in the SynDrone dataset \cite{rizzoli2023syndrone}, specific to the UAV domain, to visually illustrate the performance of several VFMs under zero-shot conditions. This example provides strong support for understanding their practical application potential.

\subsubsection{VFM for Object Detection}

The core advantage of VFMs in object detection lies in their powerful zero-shot detection capabilities. GLIP \cite{li2022grounded} unifies object detection and phrase grounding tasks, demonstrating exceptional zero-shot and few-shot transfer capabilities across various object-level recognition tasks. Zhang \emph{et al}. \cite{zhang2022dino} proposed DINO, which optimized the architecture of the DETR model \cite{carion2020end}, significantly enhancing detection performance and efficiency. Subsequent work, Grounding-DINO \cite{liu2023grounding}, introduced text supervision to improve accuracy. Additionally, DINOv2 \cite{oquab2023dinov2} adopted a discriminative self-supervised learning approach, enabling the extraction of robust image features and achieving excellent performance in downstream tasks without fine-tuning. AM-RADIO \cite{ranzinger2024radio} integrated the capabilities of VFMs such as CLIP \cite{radford2021learning}, DINOv2 \cite{oquab2023dinov2}, and SAM \cite{kirillov2023segment} through a multi-teacher distillation method, resulting in strong representational power to support complex visual tasks. DINO-WM \cite{zhou2024dino} incorporated DINOv2 into world models, enabling zero-shot planning capabilities. Additionally, YOLO-World \cite{cheng2024yolo} enhances the generalization capability of YOLO detectors through an efficient pretraining scheme, achieving outstanding performance in open vocabulary and zero-shot detection tasks.

\subsubsection{VFM for Image Segmentation}

VFMs have demonstrated significant improvements over traditional methods in image segmentation tasks. Lüdecke \emph{et al}. \cite{luddecke2022image} proposed CLIPSeg, based on the CLIP model, which supports semantic segmentation, instance segmentation, and zero-shot segmentation. Kirillov \emph{et al}. \cite{kirillov2023segment} developed the Segment Anything Model (SAM), achieving zero-shot segmentation capabilities across diverse scenarios through pretraining on large-scale and diverse datasets. Subsequent research further extended SAM's applications, such as Embodied-SAM \cite{xu2024embodiedsam} and Point-SAM \cite{zhou2024point}, which expanded SAM's functionality to 3D scenes. Open-Vocabulary SAM \cite{yuan2025open} combined SAM with CLIP's knowledge transfer strategies, effectively optimizing segmentation and recognition tasks simultaneously. Pan \emph{et al}. \cite{pan2025tokenize} proposed TAP (Tokenize Anything), a foundational model centered on visual perception, which improves the SAM architecture by introducing visual prompts to enable simultaneous completion of segmentation, recognition, and description tasks for arbitrary regions. EfficientSAM \cite{xiong2024efficientsam} and MobileSAM \cite{zhang2023faster} optimize SAM's representation, significantly reducing model complexity and achieving lightweight designs while maintaining excellent task performance. Recently, SAM2 \cite{ravi2024sam} introduced memory modules to the original model, enabling real-time segmentation for videos of arbitrary length while addressing complex challenges like occlusion and multi-object tracking. SAMURAI \cite{yang2024samurai} builds upon SAM2 by integrating a Kalman filter, addressing the limitations of memory management in SAM2, and achieving superior video segmentation performance without requiring retraining or fine-tuning.

Beyond the SAM series, other VFM architectures have also significantly advanced image segmentation. Models such as SegGPT \cite{wang2023seggpt}, Osprey \cite{yuan2024osprey}, and SEEM \cite{zou2024segment} have demonstrated notable adaptability in arbitrary segmentation tasks and multimodal scenarios. Additionally, VFMs have shown important applications in other segmentation tasks. For example, Liu \emph{et al}. \cite{liu2024segment} proposed the Seal framework for segmenting point cloud sequences, while the LISA \cite{lai2024lisa} adopted the Embedding-as-Mask approach to endow multimodal large models with reasoning-based segmentation capabilities. LISA can process complex natural language instructions and generate fine-grained segmentation results, expanding the scope and complexity of segmentation model applications.

\subsubsection{VFM for Monocular Depth Estimation (MDE)}

In the field of MDE, VFMs have also demonstrated significant technological advantages. ZoeDepth \cite{bhat2023zoedepth} achieves zero-shot depth estimation by combining relative and absolute depth estimation methods. ScaleDepth \cite{zhu2024scaledepth} decomposes depth estimation into two modules: scene scale prediction and relative depth estimation, achieving advanced performance in indoor, outdoor, unconstrained, and unseen scenarios. Additionally, Depth Anything \cite{yang2024depth1} employs many unlabeled monocular images to train an efficient and robust depth estimation method, showcasing outstanding performance in zero-shot scenarios. Depth Anything V2 \cite{yang2024depth2} introduces multiple optimizations to the original model, further improving prediction performance in complex scenes and enabling the generation of high-quality Depth images with rich details. Depth Pro \cite{bochkovskii2024depth}, based on a multi-scale ViT architecture, can quickly produce metrically accurate Depth images with high resolution and high-frequency details, making it an effective tool for handling complex depth estimation tasks.

\section{Datasets and Platforms for UAVs}

This section reviews publicly available datasets and simulation platforms relevant to UAV research, which serve as essential resources for advancing integrated studies on FM-based UAV systems. High-quality datasets form the cornerstone of UAV vision algorithms and autonomous behavior learning by offering diverse and comprehensive training data. Meanwhile, 3D simulation platforms provide safe and controlled virtual environments for the development, testing, and validation of UAV systems. These platforms can emulate complex scenarios and environmental conditions, enabling researchers to conduct experiments in a risk-free and cost-effective manner.

We present a collection of open-source datasets, primarily utilized in the development of UAV systems, all of which have been verified as publicly accessible for download. The datasets are organized in Tables \ref{tab:gdd},\ref{tab:gdd_ot},\ref{tab:gdd_ar},\ref{tab:gdd_nl},\ref{tab:dsd_t},\ref{tab:dsd_rs},\ref{tab:dsd}. The ``Year" column indicates the most recent update for each dataset; if no update has been made, the publication year of the associated paper is listed instead. The images and videos in the ``Types" column default to RGB.

The datasets cover a variety of formats, including video, RGB images (the default format in the tables), LiDAR point clouds, infrared images, Depth images, and textual data (such as descriptions or annotations). Video and RGB images are the predominant data types, while textual data is less common. Notably, some datasets have been updated to include new functionalities. For example, the EAR dataset \cite{mou2020era} was enhanced with subtitles and question-answering capabilities, evolving into the CapEAR dataset \cite{cazzato2020survey}, which is now suitable for VQA tasks. Most of the datasets listed in the tables were collected from outdoor environments and are categorized into two types: general domain datasets and domain-specific datasets.

\begin{table*}[h]
    \centering
    \caption{UAV-oriented Datasets on Environmental Perception \& Event Recognition}
    \label{tab:gdd}
    \setlength{\tabcolsep}{5pt} 
    \renewcommand{\arraystretch}{1.5} 
    \scriptsize 
    \newcolumntype{C}[1]{>{\centering\arraybackslash}m{#1}}
   \begin{tabular}{m{2cm}m{0.5cm}m{1.5cm}m{12.5cm}}
   \toprule
    \multicolumn{1}{c}{\textbf{Name}} & \multicolumn{1}{c}{\textbf{Year}} & \multicolumn{1}{c}{\textbf{Types}} & \multicolumn{1}{c}{\textbf{Description}} \\

     \midrule
    \multicolumn{4}{c}{Environmental Perception} \\
    \midrule    
    AirFisheye\cite{jaisawal2024airfisheye} & 2024 & Fisheye image \newline{}Depth image \newline{}Point cloud \newline{}IMU & Over 26,000 fisheye images in total. Data is collected at a rate of 10 frames per second. \href{https://collaborating.tuhh.de/ilt/airfisheye-dataset}{\faExternalLink}\\
    \hline
    SynDrone\cite{rizzoli2023syndrone}& 2023 & Image\newline{}Depth image\newline{}Point cloud & Contains 72,000 annotation samples, providing 28 types of pixel-level and object-level annotations. \href{https://github.com/LTTM/Syndrone}{\faExternalLink} \\    
    \hline
    WildUAV\cite{florea2021wilduav}& 2022 & Image \newline{}Video \newline{}Depth image \newline{}Metadata  & Mapping images are provided as 24-bit PNG files, with the resolution of 5280x3956. Video images are provided as JPG files at a resolution of 3840x2160. There are 16 possible class labels detailed. \href{https://github.com/hrflr/wuav}{\faExternalLink} \\

    \midrule
    \multicolumn{4}{c}{Event Recognition} \\
    \midrule 
    CapERA\cite{cazzato2020survey} & 2023 & Video \newline{} Text & 2864 videos, each with 5 descriptions, totaling 14,320 Texts. Each video lasts 5 seconds and is captured at 30 frames/second with a resolution of 640 × 640 pixels. \href{https://github.com/yakoubbazi/CapEra}{\faExternalLink}\\
    \hline
    ERA\cite{mou2020era} & 2020 & Video & A total of 2,864 videos, including disaster events, traffic accidents, sports competitions and other 25 categories. Each video is 24 frames/second for 5 seconds. \href{https://lcmou.github.io/ERA_Dataset}{\faExternalLink}\\
    \hline
    VIRAT\cite{oh2011large} & 2016 & Video & 25 hours of static ground video and 4 hours of dynamic aerial video. There are 23 event types involved. \href{https://viratdata.org/}{\faExternalLink} \\


    \bottomrule
    \end{tabular}
\end{table*}

\begin{table*}[h]
    \centering
    \caption{UAV-oriented Datasets on Object Tracking}
    \label{tab:gdd_ot}
    \setlength{\tabcolsep}{5pt} 
    \renewcommand{\arraystretch}{1.5} 
    \scriptsize 
    \newcolumntype{C}[1]{>{\centering\arraybackslash}m{#1}}
   \begin{tabular}{m{2.5cm}m{0.5cm}m{1.5cm}m{11.5cm}}
   \toprule
    \multicolumn{1}{c}{\textbf{Name}} & \multicolumn{1}{c}{\textbf{Year}} & \multicolumn{1}{c}{\textbf{Types}} & \multicolumn{1}{c}{\textbf{Description}} \\
    \midrule

    WebUAV-3M\cite{zhang2022webuav} & 2024 & Video \newline{}Text \newline{}Audio & 4,500 videos totaling more than 3.3 million frames with 223 target categories, providing natural language and audio descriptions. \href{https://github.com/983632847/WebUAV-3M}{\faExternalLink}\\ 
    \hline
    UAVDark135
    \cite{li2022all}& 2022 & Video & 135 video sequences with over 125,000 manually annotated frames. \href{https://vision4robotics.github.io/project/uavdark135/}{\faExternalLink}\\
    \hline
    DUT-VTUAV\cite{zhang2022visible}& 2022 & RGB-T Image & Nearly 1.7 million well-aligned visible-thermal (RGB-T) image pairs with 500 sequences for unveiling the power of RGB-T tracking. Including 13 sub-classes and 15 scenes across 2 cities. \href{https://github.com/zhang-pengyu/DUT-VTUAV}{\faExternalLink}\\
    \hline
    TNL2K\cite{wang2021towards}& 2022 & Video \newline{}Infrared video \newline{}Text& 2,000 video sequences, comprising 1,244,340 frames and 663 words. \href{https://github.com/wangxiao5791509/TNL2K_evaluation_toolkit}{\faExternalLink}\\
    \hline
    PRAI-1581\cite{zhang2020person}& 2020 & Image & 39,461 images of 1581 person identities. \href{https://github.com/stormyoung/PRAI-1581}{\faExternalLink}\\
    \hline
    VOT-ST2020/VOT-RT2020\cite{kristan2020eighth} & 2020 & Video & 1,000 sequences, each varying in length, with an average length of approximately 100 frames. \href{https://votchallenge.net/vot2020/dataset.html}{\faExternalLink}\\
    \hline
    VOT-LT2020\cite{kristan2020eighth}& 2020 & Video & 50 sequences, each with a length of approximately 40,000 frames. \href{https://votchallenge.net/vot2020/dataset.html}{\faExternalLink}\\
    \hline
    VOT-RGBT2020\cite{kristan2020eighth} & 2020 & Video \newline{}Infrared video & 50 sequences, each with a length of approximately 40,000 frames. \href{https://votchallenge.net/vot2020/dataset.html}{\faExternalLink}\\
    \hline
    VOT-RGBD2020\cite{kristan2020eighth} & 2020 & Video \newline{}Depth image & 80 sequences with a total of approximately 101,956 frames. \href{https://votchallenge.net/vot2020/dataset.html}{\faExternalLink}\\
    \hline
    GOT-10K\cite{huang2019got}& 2019 & Image \newline{}Video & 420 video clips belonging to 84 object categories and 31 motion categories. \href{http://got-10k.aitestunion.com/}{\faExternalLink}\\
    \hline
    DTB70\cite{li2017visual}& 2017 & Video & 70 video sequences, each consisting of multiple video frames, with each frame containing an RGB image at a resolution of 1280x720 pixels. \href{https://github.com/flyers/drone-tracking}{\faExternalLink}\\
    \hline
    Stanford Drone\cite{robicquet2016learning}& 2016 & Video & 19,000 + target tracks, containing 6 types of targets, about 20,000 target interactions, 40,000 target interactions with the environment, covering 100 + scenes in the university campus. \href{https://cvgl.stanford.edu/projects/uav_data/}{\faExternalLink} \\
    \hline
    COWC\cite{mundhenk2016large} & 2016 & Image & 32,716 unique vehicles and 58,247 non-vehicle targets were labeled. Covering 6 different geographical areas. \href{https://gdo152.llnl.gov/cowc/}{\faExternalLink} \\

    \bottomrule
    \end{tabular}
\end{table*}

\subsection{General Domain Datasets}
General domain datasets are designed to cater to a wide range of scenarios and are further categorized based on specific tasks, including Environmental Perception, Event Recognition, Object Tracking, Action Recognition, and Navigation. Within the Environmental Perception category, we focus on tasks such as object detection, segmentation, and depth estimation. Although tasks like event recognition, object tracking, and action recognition can also be considered part of Environmental Perception, we have listed them separately to provide a clearer presentation of the datasets.

\subsubsection{Environmental Perception}
This part presents the datasets used primarily for object detection, segmentation, and depth estimation, as shown in Table \ref{tab:gdd}. For instance, the AirFisheye dataset \cite{jaisawal2024airfisheye} is specifically designed for tasks such as object detection, segmentation, and depth estimation in complex urban environments captured by UAVs. Its multimodal data, including visual, thermal imaging, and LiDAR, provide comprehensive information for analyzing scenes in these challenging urban settings. The SynDrone dataset \cite{rizzoli2023syndrone} is a large-scale synthetic dataset generated using Carla, intended for detection and segmentation tasks in urban environments. The WildUAV dataset \cite{florea2021wilduav} provides high-resolution RGB images and depth ground truth data, focusing on monocular visual depth estimation while supporting precise drone flight control in complex environments.

\subsubsection{Event Recognition}
The typical datasets used for event recognition are listed in Table \ref{tab:gdd}. The EAR dataset \cite{mou2020era} serves as a video-based benchmark for event recognition, encompassing 25 event classes, including post-earthquake, flood, fire, landslide, mudslide, traffic collision, traffic congestion, harvesting, plowing, construction, police chase, conflict, various sports (e.g., baseball, basketball, cycling), and social activities (e.g., parties, concerts, protests, religious activities). It consists of 2,864 videos, each lasting 5 seconds,  collected from YouTube using ``drone" and ``UAV" as search keywords. Similarly, the VIRAT dataset \cite{oh2011large} focuses on event recognition in surveillance videos, including events like traffic accidents and crowd gatherings. Although not captured by UAVs, the VIRAT dataset offers similar aerial perspectives, making it relevant for drone-based scene analysis. Together, these datasets provide valuable resources for advancing research in event detection and scene understanding, particularly in the context of integrating UAV applications with LLMs.

\subsubsection{Object Tracking}

Object tracking tasks rely on diverse datasets to advance research across various domains. Table \ref{tab:gdd_ot} lists typical datasets for this task. The WebUAV-3M dataset \cite{zhang2022webuav} is a large-scale benchmark for UAV object tracking, comprising 4,500 videos across 233 object categories. It serves as a solid foundation for UAV tracking in general scenarios and includes multimodal data, such as audio and natural language descriptions, enabling the exploration of multimodal UAV tracking approaches. The TNL2K dataset \cite{wang2021towards} focuses on natural language-guided object tracking and includes 2,000 video sequences annotated with bounding boxes and detailed natural language descriptions that capture the target object's category, shape, attributes, features, and spatial location. To address challenging scenarios, TNL2K incorporates adversarial samples and sequences with significant appearance changes, offering both RGB and infrared modalities to support cross-modal tracking research. The VOT2020 dataset \cite{kristan2020eighth} provides a comprehensive collection of five specialized datasets tailored to specific tasks: short-term tracking, real-time tracking, long-term tracking, thermal tracking, and depth tracking. These datasets collectively address a wide range of tracking challenges, fostering innovation across different tracking paradigms.

\subsubsection{Action Recognition}
Enabling drones to comprehend human actions and interpret commands via gestures is a pivotal area of research. Table \ref{tab:gdd_ar} lists UAV-oriented datasets for action recognition. The Aeriform In-Action dataset \cite{kapoor2023aeriform} targets human action recognition in aerial videos, featuring 32 high-resolution videos across 13 action categories. This dataset is specifically designed to address the unique challenges associated with action recognition in aerial surveillance. The MEVA dataset \cite{corona2021meva} offers a large-scale, multi-view, multimodal dataset comprising 9,300 hours of continuous video captured by UAVs and ground cameras. It covers 37 activity categories and facilitates advanced tasks, such as multi-view activity detection. Additionally, the UAV-Human dataset \cite{li2021uav} provides 67,428 multimodal video sequences, encompassing 119 subjects for action recognition. In addition to action recognition, it supports tasks such as pose estimation and person re-identification. With its diverse range of backgrounds, lighting conditions, and environments, the dataset serves as a comprehensive benchmark for drone-based human behavior analysis.

\begin{table*}[h]
    \centering
    \caption{UAV-oriented Datasets on Action Recognition}
    \label{tab:gdd_ar}
    \setlength{\tabcolsep}{5pt} 
    \renewcommand{\arraystretch}{1.5} 
    \scriptsize 
    \newcolumntype{C}[1]{>{\centering\arraybackslash}m{#1}}
   \begin{tabular}{m{2.5cm}m{0.5cm}m{1.5cm}m{11.5cm}}
   \toprule
    \multicolumn{1}{c}{\textbf{Name}} & \multicolumn{1}{c}{\textbf{Year}} & \multicolumn{1}{c}{\textbf{Types}} & \multicolumn{1}{c}{\textbf{Description}} \\
    \hline
    Aeriform in-action\cite{kapoor2023aeriform}& 2023 & Video & 32 videos, 13 types of action, 55,477 frames, 40,000 callouts. \href{https://surbhi-31.github.io/Aeriform-in-action/}{\faExternalLink}\\
    \hline
    MEVA\cite{corona2021meva} & 2021 & Video \newline{}Infrared video \newline{} GPS \newline{}Point cloud & Total 9,300 hours of video, 144 hours of activity notes, 37 activity types, over 2.7 million GPS track points. \href{https://mevadata.org/}{\faExternalLink}\\
    \hline
    UAV-Human\cite{li2021uav}& 2021 & Video \newline{}Night-vision video \newline{}Fisheye video \newline{}Depth video \newline{}Infrared video \newline{}Skeleton & 67,428 videos (155 types of actions, 119 subjects), 22,476 frames of annotated key points (17 key points), 41,290 frames of people re-recognition (1,144 identities), 22,263 frames of attribute recognition (such as gender, hat, backpack, etc.). \href{https://github.com/SUTDCV/UAV-Human}{\faExternalLink} \\
    \hline
    MOD20\cite{perera2020multiviewpoint}& 2020 & Video & 20 types of action, 2,324 videos, 503,086 frames. \href{https://asankagp.github.io/mod20/}{\faExternalLink}\\
    \hline
    NEC-DRONE\cite{choi2020unsupervised}& 2020 & Video & 5,250 videos containing 256 minutes of action videos involving 19 actors and 16 action categories \href{https://www.nec-labs.com/research/media-analytics/projects/unsupervised-semi-supervised-domain-adaptation-for-action-recognition-from-drones/}{\faExternalLink}\\    
    \hline
    Drone-Action\cite{perera2019drone} & 2019 & Video & 240 HD videos, 66,919 frames, 13 types of action. \href{https://asankagp.github.io/droneaction/}{\faExternalLink}\\
    \hline
    UAV-GESTURE\cite{perera2018uav} & 2019 & Video & 119 videos, 37,151 frames, 13 types of gestures, 10 actors. \href{https://asankagp.github.io/uavgesture/}{\faExternalLink}\\

           \bottomrule
    \end{tabular}
\end{table*}

\subsubsection{Navigation and Localization}

Table \ref{tab:gdd_nl} presents UAV-oriented datasets for navigation and localization. The CityNav dataset \cite{lee2024citynav} is a dataset designed for language-guided aerial navigation tasks, aimed at assisting drones in navigating city-scale 3D environments using natural language instructions. The dataset comprises 32,000 tasks, offering extensive geographic information and detailed urban environment models. The AerialVLN dataset \cite{liu2023aerialvln} focuses on drone navigation through the integration of visual and linguistic cues, enabling drones to perform flight tasks in complex environments based on natural language commands, thereby enhancing their adaptability in dynamic settings. The VIGOR dataset \cite{zhu2021vigor} provides a cross-view image localization dataset that facilitates accurate geographical positioning of drones from diverse perspectives, improving image matching and position calibration accuracy in complex geographical environments. The University-1652 dataset \cite{zheng2020university} serves as a benchmark for cross-view geo-localization, bridging the visual gap between ground-level and satellite perspectives by incorporating drone-view images. It includes paired images from synthetic drones, satellites, and ground cameras for 1,652 universities, supporting two tasks: drone-view target localization and drone navigation.



\begin{table*}[h]
    \centering
    \caption{UAV-oriented Datasets on Navigation and Localization}
    \label{tab:gdd_nl}
    \setlength{\tabcolsep}{5pt} 
    \renewcommand{\arraystretch}{1.5} 
    \scriptsize 
    \newcolumntype{C}[1]{>{\centering\arraybackslash}m{#1}}
   \begin{tabular}{m{2.5cm}m{0.5cm}m{1cm}m{12cm}}
   \toprule
    \multicolumn{1}{c}{\textbf{Name}} & \multicolumn{1}{c}{\textbf{Year}} & \multicolumn{1}{c}{\textbf{Types}} & \multicolumn{1}{c}{\textbf{Description}} \\
    \hline
    CityNav\cite{lee2024citynav}& 2024 & Image \newline{}Text & 32,000 natural language descriptions and companion tracks. \href{https://water-cookie.github.io/city-nav-proj/}{\faExternalLink} \\
    \hline
    CNER-UAV\cite{yao2024can}& 2024 & Text & 12,000 labeled samples containing 5 types of address labels (e.g., building, unit, floor, room, etc.). \href{https://github.com/zhhvvv/CNER-UAV}{\faExternalLink}\\
    \hline
    AerialVLN\cite{liu2023aerialvln} & 2023 & Path\newline{}Text & It contains 25 city-level scenes, including urban areas, factories, parks and villages. A total of 8,446 paths. Each path is provided with 3 natural language descriptions, totaling 25,338 instructions. 
    \href{https://github.com/AirVLN/AirVLN}{\faExternalLink}\\
    \hline
    DenseUAV\cite{dai2023vision}& 2023 & Image & Training set: 6,768 UAV images, 13,536 satellite images. Test set: 2,331 UAV query images and 4,662 satellite images. \href{https://github.com/Dmmm1997/DenseUAV}{\faExternalLink}\\
    \hline
    map2seq\cite{schumann2022analyzing} & 2022 & Image \newline{}Text \newline{}Map path & 29,641 panoramic images, 7,672 navigation instruction Texts. \href{https://map2seq.schumann.pub/dataset/download/}{\faExternalLink} \\
    \hline
    VIGOR \cite{zhu2021vigor}& 2021 & Image & 90,618 aerial images, 238,696 street panorama. \href{https://github.com/Jeff-Zilence/VIGOR}{\faExternalLink}\\
    \hline
    University-1652\cite{zheng2020university}& 2020 & Image & 1,652 university buildings, involving 72 universities, 50,218 training images, 37,855 UAV perspective query images, 701 satellite perspective query images, and an additional 21,099 ordinary perspective and 5,580 street view perspective images were collected for training. \href{https://github.com/layumi/University1652-Baseline}{\faExternalLink}\\ 
    
               \bottomrule
    \end{tabular}
\end{table*}

\subsection{Domain-specific Datasets}
Compared to general-domain datasets, domain-specific datasets are tailored for particular applications and categorized according to the specific domains they address, including Transportation, Remote Sensing, Agriculture, Industrial Applications, Emergency Response, Military Operations, and Wildlife.

\subsubsection{Transportation}
Transportation scenes are among the most prevalent scenarios in UAV datasets and this part highlights datasets (as shown in Table \ref{tab:dsd_t}) specifically designed for traffic monitoring, as well as vehicle and pedestrian detection tasks, which are key applications of UAV technology. The TrafficNight dataset \cite{zhang2025trafficnight} is an aerial multimodal dataset for nighttime vehicle monitoring, designed to address the limitations of existing aerial datasets in terms of lighting conditions and vehicle type representativeness. The dataset combines vertical RGB and thermal infrared imaging technologies, covering various scenes, including those with numerous semi-trailers, and provides specialized annotations. It also includes corresponding HD-MAP data for multi-vehicle tracking. The VisDrone dataset \cite{zhu2021detection} is a large-scale benchmark supporting object detection and both single- and multi-object tracking in images and videos. Collected from 14 cities across China, it features high diversity and challenging scenarios, making it well-suited for evaluating algorithms in complex urban and suburban environments. The CADP dataset \cite{shah2018cadp} emphasizes traffic accident analysis, enhancing small-object detection accuracy (e.g., pedestrians) using CCTV traffic monitoring videos. It integrates contextual mining techniques and an LSTM-based architecture for accident prediction. The CARPK dataset \cite{hsieh2017drone} introduces a novel method for parking lot vehicle counting using a spatially regularized region proposal network, called Layout Proposal Network. It includes high-resolution UAV imagery with over 90,000 vehicles, enhancing object detection and counting performance. The iSAID dataset \cite{waqas2019isaid} offers high-quality annotations for instance segmentation tasks, encompassing 655,451 labeled instances across 15 categories, thus supporting accurate object detection and scene analysis in UAV applications. Collectively, these datasets advance research in vehicle detection, object tracking, traffic monitoring, and UAV autonomous navigation, offering robust data resources for applications in intelligent transportation, UAV-based patrols, and delivery systems.

\begin{table*}[h]
    \centering
    \caption{UAV-oriented Datasets on Transportation}
    \label{tab:dsd_t}
    \setlength{\tabcolsep}{5pt} 
    \renewcommand{\arraystretch}{1.5} 
    \scriptsize 
    \newcolumntype{C}[1]{>{\centering\arraybackslash}m{#1}}
   \begin{tabular}{m{1.5cm}m{0.5cm}m{1.5cm}m{12.5cm}}
   \toprule
    \multicolumn{1}{c}{\textbf{Name}} & \multicolumn{1}{c}{\textbf{Year}} & \multicolumn{1}{c}{\textbf{Types}} & \multicolumn{1}{c}{\textbf{Description}} \\
    \hline

    TrafficNight\cite{zhang2025trafficnight} & 2024 & Image \newline{}Infrared Image \newline{}Video \newline{}Infrared Video \newline{}Map & The dataset consists of 2,200 pairs of annotated thermal infrared and RGB image data, as well as video data from 7 traffic scenes, with a total duration of approximately 240 minutes. Each scene includes a high-precision map, providing a detailed layout and topological information. \href{https://github.com/AIMSPolyU/TrafficNight}{\faExternalLink}\\
    \hline
    
    VisDrone\cite{zhu2021detection}& 2022 & mage\newline{} Video& 263 videos, 179,264 frames. 10,209 still images. More than 2,500,000 object instance annotations. The data covers 14 different cities, covering a wide range of weather and light conditions. \href{http://aiskyeye.com/home/}{\faExternalLink}\\
    \hline
    ITCVD\cite{yang2018deep}& 2020 & Image & A total of 173 aerial images were collected, including 135 in the training set with 23,543 vehicles and 38 in the test set with 5,545 vehicles. There is 60\% regional overlap between the images, and there is no overlap between the training set and the test set. \href{https://research.utwente.nl/en/datasets/itcvd-dataset}{\faExternalLink}\\
    \hline
    UAVid\cite{lyu2020uavid}& 2020 & Image \newline{}Video & 30 videos, 300 images, 8 semantic category annotations. \href{https://uavid.nl/}{\faExternalLink} \\
    \hline
    AU-AIR\cite{bozcan2020air}& 2020 & Video \newline{}GPS \newline{}Altitude \newline{}IMU \newline{}Speed & 32,823 frames of video, 1920x1080 resolution, 30 FPS, divided into 30,000 training validation samples and 2,823 test samples. The total duration of the 8 videos is about 2 hours, with a total of 132,034 instances, distributed in 8 categories.  \href{https://bozcani.github.io/auairdataset}{\faExternalLink}\\
    \hline
    iSAID\cite{waqas2019isaid}& 2020 & Image & Total images: 2,806. Total number of instances: 655,451. Test set: 935 images (not publicly labeled, used to evaluate the server). \href{https://captain-whu.github.io/iSAID/}{\faExternalLink} \\
    \hline
    CARPK\cite{hsieh2017drone}& 2018 & Image & 1448 images, approx. 89,777 vehicles, providing box annotations. \href{https://lafi.github.io/LPN/}{\faExternalLink} \\
    \hline
    highD\cite{krajewski2018highd}& 2018 & Video \newline{}Trajectory & 16.5 hours, 110,000 vehicles, 5,600 lane changes, 45,000 km, totaling approximately 447 hours of vehicle travel data; 4 predefined driving behavior labels. \href{https://levelxdata.com/highd-dataset/}{\faExternalLink} \\
    \hline
    UAVDT\cite{du2018unmanned}& 2018 & Video  \newline{}Weather \newline{}Altitude \newline{}Camera angle & 100 videos, about 80,000 frames, 30 frames per second, containing 841,500 target boxes, covering 2,700 targets. \href{https://sites.google.com/view/grli-uavdt/}{\faExternalLink}\\
    \hline
    CADP\cite{shah2018cadp} & 2016 & Video & A total of 5.24 hours, 1,416 traffic accident clips, 205 full-time and space annotation videos.  \href{https://ankitshah009.github.io/accident_forecasting_traffic_camera}{\faExternalLink}\\
    \hline
    VEDAI\cite{razakarivony2016vehicle}& 2016 & Image & 1,210 images (1024 × 1024 and 512 × 512 pixels), 9 types of vehicles, containing about 6,650 targets in total. \href{https://downloads.greyc.fr/vedai}{\faExternalLink}\\
    
           \bottomrule
    \end{tabular}
\end{table*}

\subsubsection{Remote Sensing}
In the field of remote sensing, several innovative datasets, as shown in Table \ref{tab:dsd_rs}, provide substantial support for tasks such as object detection, classification, localization, and image analysis \cite{zhang2023aerial}. The xView dataset \cite{lam2018xview} is a large-scale satellite image dataset that containing over one million annotations, across multiple object categories, making it particularly suitable for object detection and image segmentation tasks, especially in complex backgrounds and challenging environments. The DOTA dataset \cite{xia2018dota} focuses on object detection in high-resolution aerial images, covering multiple object categories such as aircraft, vehicles, and buildings, and is suitable for multi-object detection and classification tasks in complex scenes. The RSICD dataset \cite{lu2017exploring} is primarily used for scene classification tasks in remote sensing images and supports language description generation, providing a standardized benchmark that promotes research in image understanding and automated annotation techniques. 
 RemoteCLIP \cite{liu2024remoteclip} introduces a remote sensing visual-language model that enhances semantic analysis and image retrieval of remote sensing images through self-supervised learning and masked image modeling, advancing the application of drones in remote sensing data analysis.

\begin{table*}[h]
    \centering
    \caption{UAV-oriented Datasets on Remote Sensing}
    \label{tab:dsd_rs}
    \setlength{\tabcolsep}{5pt} 
    \renewcommand{\arraystretch}{1.5} 
    \scriptsize 
    \newcolumntype{C}[1]{>{\centering\arraybackslash}m{#1}}
   \begin{tabular}{m{2.5cm}m{0.5cm}m{0.8cm}m{12.2cm}}
   \toprule
    \multicolumn{1}{c}{\textbf{Name}} & \multicolumn{1}{c}{\textbf{Year}} & \multicolumn{1}{c}{\textbf{Types}} & \multicolumn{1}{c}{\textbf{Description}} \\
    \hline

    RET-3\cite{liu2024remoteclip} & 2024 & Image \newline{}Text & Approximately 13,000 samples. Including RSICD, RSITMD and UCM. \href{https://github.com/ChenDelong1999/RemoteCLIP?utm_source=chatgpt.com}{\faExternalLink} \\
    \hline
    DET-10\cite{liu2024remoteclip} & 2024 & Image & In the object detection dataset, the number of objects per image ranges from 1 to 70, totaling about 80,000 samples. \href{https://github.com/ChenDelong1999/RemoteCLIP?utm_source=chatgpt.com}{\faExternalLink}\\
    \hline
    SEG-4\cite{liu2024remoteclip} & 2024 & Image & The segmented data set covers different regions and resolutions, totaling about 72,000 samples. \href{https://github.com/ChenDelong1999/RemoteCLIP?utm_source=chatgpt.com}{\faExternalLink}\\
    \hline
    DIOR\cite{li2020object} & 2020 & Image & 23,463 images, containing 192,472 target instances, covering 20 categories, including aircraft, vehicles, ships, bridges, etc., each category contains about 1,200 instances. \href{http://www.escience.cn/people/gongcheng/DIOR.html}{\faExternalLink} \\ 
    \hline
    TGRS-HRRSD\cite{zhang2019hierarchical} & 2019 & Image & Total images: 21,761. 13 categories, including aircraft, vehicles, bridges, etc. The total number of targets is approximately 53,000 targets. \href{https://github.com/CrazyStoneonRoad/TGRS-HRRSD-Dataset}{\faExternalLink} \\
    \hline
    xView\cite{lam2018xview} & 2018 & Image & There are more than 1 million goals and 60 categories, including vehicles, buildings, facilities, boats and so on, which are divided into seven parent categories and several sub-categories. \href{https://xviewdataset.org/}{\faExternalLink} \\
    \hline
    DOTA\cite{xia2018dota} & 2018 & Image & 2806 images, 188, 282 targets, 15 categories.  \href{https://captain-whu.github.io/DOTA/}{\faExternalLink}  \\
    \hline
    RSICD\cite{lu2017exploring} & 2018 & Image \newline{}Text & 10,921 images, 54,605 descriptive sentences. \href{https://github.com/201528014227051/RSICD_optimal}{\faExternalLink} \\       
    \hline
    HRSC2016\cite{liu2017high} & 2017 & Image & 3,433 instances, totaling 1,061 images, including 70 pure ocean images and 991 images containing mixed land-sea areas. 2,876 marked vessel targets. 610 unlabeled images. \href{http://www.escience.cn/people/liuzikun/DataSet.html}{\faExternalLink} \\
    \hline
    RSOD\cite{long2017accurate} & 2017 & Image & Contains 4 types of targets (tank, aircraft, overpass, playground) with 12,000 positive samples and 48,000 negative samples. \href{https://github.com/RSIA-LIESMARS-WHU/RSOD-Dataset-}{\faExternalLink} \\
    \hline
    NWPU-RESISC45\cite{cheng2017remote}
    & 2017 & Image & A total of 31,500 images, covering 45 scene categories, 700 images per category, resolution 256 × 256 pixels, spatial resolution from 0.2m to 30m. \href{http://pan.baidu.com/s/1mifR6tU}{\faExternalLink}  \\
    \hline
    NWPU VHR-10\cite{cheng2014multi} & 2014 & Image & 800 high-resolution images, of which 650 contain targets and 150 are background images, covering 10 categories (such as aircraft, ships, bridges, etc.), totaling more than 3,000 targets. \href{https://github.com/Gaoshuaikun/NWPU-VHR-10}{\faExternalLink} \\
               \bottomrule
    \end{tabular}
\end{table*}

 \subsubsection{Agriculture}
The agricultural section summarizes only publicly available datasets from the past two years, as shown in Table \ref{tab:dsd}, as several reviews have covered datasets before 2023. Agricultural datasets are commonly used for object detection to identify weeds, invasive plants, or plant diseases and pests, while semantic segmentation is often used for field division. The Avo-AirDB dataset \cite{amraoui2022avo} is specifically designed for agricultural image segmentation and classification, providing high-resolution images of avocado crops to support plant identification and health monitoring in precision agriculture. The CoFly-WeedDB dataset \cite{krestenitis2022cofly} consists of 201 aerial images, capturing three types of weeds that interfere with cotton crops, along with corresponding annotated images. The WEED-2C Dataset \cite{tetila2024real} focuses on training UAV images for species detection of weeds in soybean fields, automatically identifying two weed species.

\subsubsection{Industry}
Using drone imagery for industrial inspections, particularly in infrastructure maintenance, has become increasingly important. Table \ref{tab:dsd} lists several typical datasets. The UAPD dataset \cite{zhong2022multi} focuses on detecting asphalt pavement cracks through drone imagery and the YOLO architecture, aiming to enhance road and highway maintenance efficiency via automated crack detection. The InsPLAD dataset \cite{vieira2023insplad} is specifically designed for power line asset detection, containing drone images focused on infrastructure such as power lines, towers, and insulators. By providing images under diverse environmental conditions, this dataset supports the development of automated inspection systems to identify damage or aging in power equipment, thereby improving the efficiency and accuracy of power line inspections.

\subsubsection{Emergency Response}
These datasets as shown in Table \ref{tab:dsd} are typically used to enhance the visual understanding capabilities of drones in disaster rescue scenarios, particularly in post-disaster scene analysis, disaster area monitoring, environmental assessment, and rescue operations. They facilitate rapid image recognition, object detection, and scene understanding tasks.  The dataset \cite{mishra2020drone} proposed by Mishra \emph{et al}. explores drone applications in natural disaster monitoring and search-and-rescue operations, highlighting the potential for rapid deployment and autonomous management of drones in disaster zones. The AFID dataset \cite{wang2023aerial} provides aerial imagery for water channel surveillance and disaster early warning, supporting the training of deep semantic segmentation models. The FloodNet dataset \cite{rahnemoonfar2021floodnet} offers high-resolution aerial imagery for post-disaster scene understanding, primarily intended to assist in post-disaster assessments and emergency rescue operations. By utilizing these datasets, researchers can significantly improve image analysis capabilities in disaster response and advance the practical application of drone technology in disaster rescue efforts.

\subsubsection{Military}
The MOCO dataset \cite{pan2024military} is designed for the Military Image Captioning (MilitIC) task, which focuses on generating textual intelligence from images captured by low-altitude UAVs and UGVs (Unmanned Ground Vehicles) in military contexts. The dataset includes a training set with 7,192 images and 35,960 captions, as well as a test set with 257 images and 1,285 captions. MilitIC, as a vision-language learning task, aims to automatically generate descriptive captions for military images, thereby enhancing situational awareness and supporting decision-making. By integrating image data with textual descriptions, this approach improves intelligence capabilities and operational efficiency in the military domain.

\subsubsection{Wildlife}
The WAID \cite{mou2023waid} is a large-scale, multi-class, high-quality dataset specifically designed to support the use of drones in wildlife monitoring. The dataset includes 14,375 drone images captured under various environmental conditions, covering six species of wildlife and multiple types of habitats.

\begin{table*}[h]
    \centering
    \caption{UAV-oriented Datasets on Agriculture \& Industry \& Emergency Response \& Military \& Wildlife}
    \label{tab:dsd}
    \setlength{\tabcolsep}{5pt} 
    \renewcommand{\arraystretch}{1.5} 
    \scriptsize 
    \newcolumntype{C}[1]{>{\centering\arraybackslash}m{#1}}
   \begin{tabular}{m{2.5cm}m{0.5cm}m{1cm}m{12cm}}
   \toprule
    \multicolumn{1}{c}{\textbf{Name}} & \multicolumn{1}{c}{\textbf{Year}} & \multicolumn{1}{c}{\textbf{Types}} & \multicolumn{1}{c}{\textbf{Description}} \\
    
    \midrule
    \multicolumn{4}{c}{Agriculture} \\
    \midrule
    WEED-2C\cite{tetila2024real}& 2024 & Image & Contains 4,129 labeled samples covering 2 weed species. \href{https://github.com/EvertonTetila/WEED2C-Dataset/?tab=readme-ov-file}{\faExternalLink} \\
    \hline
    CoFly-WeedDB\cite{krestenitis2022cofly} & 2023 & Image \newline{}Health data & Consisting of 201 aerial images, different weed types of 3 disturbed row crops (cotton) and their corresponding annotated images were captured. \href{https://github.com/CoFly-Project/CoFly-WeedDB/blob/main/README.md?utm_source=chatgpt.com}{\faExternalLink} \\    
    \hline
    Avo-AirDB\cite{amraoui2022avo}& 2022 & Image & 984 high-resolution RGB images (5472 × 3648 pixels), 93 of which have detailed polygonal annotations, divided into 3 to 4 categories (small, medium, large, and background). \href{https://github.com/LCSkhalid/Avo-AirDB}{\faExternalLink} \\
    
    \midrule
    \multicolumn{4}{c}{Industry} \\
    \midrule
    UAPD\cite{zhong2022multi} & 2021 & Image & There are 2,401 crack images in the original data and 4,479 crack images after data enhancement. \href{https://github.com/tantantetetao/UAPD-Pavement-Distress-Dataset}{\faExternalLink}\\
    \hline
    InsPLAD\cite{vieira2023insplad}& 2023 & Image & 10,607 UAV images containing 17 classes of power assets with a total of 28,933 labeled instances, and defect labels for 5 assets with a total of 402 defect samples classified into 6 defect types. \href{https://github.com/andreluizbvs/InsPLAD/}{\faExternalLink} \\

    \midrule
    \multicolumn{4}{c}{Emergency Response} \\
    \midrule
    AFID\cite{wang2023aerial}& 2023 & Image & A total of 816 images with resolutions of 2720 × 1536 and 2560 × 1440. Contains 8 semantic segmentation categories. \href{https://purr.purdue.edu/publications/4105/1}{\faExternalLink}\\  
    \hline
    FloodNet\cite{rahnemoonfar2021floodnet}& 2021 & Image \newline{}Text & The whole dataset has 2,343 images, divided into training (~60\%), validation (~20\%), and test (~20\%) sets. The semantic segmentation labels include: Background, Building Flooded, Building Non-Flooded, Road Flooded, Road Non-Flooded, Water, Tree, Vehicle, Pool, Grass. \href{https://github.com/BinaLab/FloodNet-Supervised_v1.0}{\faExternalLink}\\
    \hline
    Mishra \emph{et al}. \cite{mishra2020drone}& 2020 & Image & 2,000 images with 30,000 action instances covering multiple human behaviors. \href{https://www.leadingindia.ai/data-set}{\faExternalLink} \\

    \midrule
    \multicolumn{4}{c}{Military} \\
    \midrule
    MOCO \cite{pan2024military}& 2024 & Image\newline{}Text & 7,449 images, 37,245 captions. \href{https://github.com/Panlizhi/MOCO}{\faExternalLink} \\

    \midrule
    \multicolumn{4}{c}{Wildlife} \\
    \midrule
    WAID \cite{mou2023waid}& 2023 & Image & 14,375 UAV images covering 6 species of wildlife and multiple environment types. \href{https://github.com/xiaohuicui/WAID}{\faExternalLink}\\ 
               \bottomrule
    \end{tabular}
\end{table*}

\subsection{3D Simulation Platforms}

The 3D simulation platform plays a crucial role in the development and application of drones by providing safe, controllable, and diverse testing scenarios for intelligent drone training within highly simulated virtual environments. These environments encompass complex conditions such as varying weather, lighting, wind speed, terrain, and obstacles. Such platforms are capable of generating large-scale, accurately labeled multimodal datasets for training and validation. Additionally, the simulation platform supports the modeling of multi-machine collaborative tasks, assessing drones' collaborative capabilities, communication, and collision avoidance strategies within shared spaces. This effectively reduces risks and costs associated with real-world testing. Hardware-in-the-loop (HIL) simulation further integrates virtual testing with real hardware, helping identify potential issues and verify system reliability. In summary, 3D simulation platforms are pivotal in intelligent training, dataset generation, collaborative task execution, and hardware verification, significantly accelerating the development and deployment of drone technology.

\subsubsection{AirSim}

AirSim \cite{shah2018airsim} is an open-source, cross-platform simulator developed by Microsoft, designed specifically for the research and development of drones, autonomous vehicles, and other autonomous systems. Built on the Unreal Engine, the platform offers highly realistic physical simulation environments and visual effects, allowing users to test and validate the performance of algorithms in virtual scenarios. AirSim supports the simulation of various devices and sensors, including cameras, LiDAR, IMUs, GPS, and more, while providing comprehensive control over the environment and vehicles through its powerful API. Developers can extend the platform using Python and C++, enabling integration of cutting-edge technologies from fields such as machine learning, computer vision, and robotics. In addition to simulating drones and ground vehicles, the platform can model complex dynamic scenarios, including weather changes, collision detection, and physical interactions, helping users accelerate prototype validation and algorithm optimization in a safe and controlled virtual environment.

\subsubsection{Carla}

CARLA \cite{dosovitskiy2017carla} is an open-source autonomous driving simulation platform built on Unreal Engine, widely used for the development, training, and validation of algorithms for intelligent systems. Its highly realistic simulation environment supports complex urban scenarios, including road networks, dynamic traffic, pedestrian behavior, and diverse weather and lighting conditions, providing a virtual testing ground for perception, localization, planning, and control algorithms. CARLA supports the simulation of various sensors such as cameras, LiDAR, radar, IMUs, and GPS, and allows users to access its Python or C++ APIs, as well as interfaces supporting ROS, enabling researchers to quickly develop and test algorithms for navigation, obstacle avoidance, path planning, and environmental perception. Additionally, CARLA offers data recording and playback functionalities, supports multi-agent tasks, and integrates reinforcement learning applications, providing a safe, efficient, and repeatable testing platform for algorithm development in scenarios such as low-altitude logistics, monitoring, and patrolling for UAVs.

\subsubsection{NVIDIA Isaac Sim}

NVIDIA Isaac Sim \cite{IsaacSim} is a physics-based robotic simulation platform built on the NVIDIA Omniverse platform, providing a high-precision virtual environment for the development, testing, and validation of robots and autonomous systems. The platform leverages NVIDIA's powerful GPU acceleration and physics engine technologies, including PhysX and RTX real-time rendering, to present highly realistic simulation scenes with accurate physical interactions, lighting effects, and multi-sensor data generation. Isaac Sim offers a wide range of tools and plugins, allowing integration with various robotic frameworks, and supports the full development process from perception and motion planning to control algorithms. In addition to its applications in traditional robotics, Isaac Sim can be extended to the UAV domain, supporting drone navigation, obstacle avoidance, target tracking, and multi-agent collaboration tasks through its flexible environmental configuration, sensor simulation (including cameras, LiDAR, IMUs, and GPS), and complex dynamics modeling. The platform combines simulation reinforcement learning capabilities, data collection features, and digital twin support for real-world scenarios, enabling accelerated algorithm development for UAVs in areas such as logistics, environmental monitoring, and disaster response, while providing researchers and developers with an efficient, safe, and scalable testing environment.

\subsubsection{AerialVLN Simulator}

The AerialVLN Simulator \cite{liu2023aerialvln} is a high-fidelity virtual simulation platform designed specifically for research on drone agents. It integrates Unreal Engine 4 and Microsoft AirSim technologies to realistically simulate typical 3D urban environments, including cities such as Shanghai, Shenzhen, campuses, and residential areas, with coverage ranging from 30 hectares to 3,700 hectares. The platform supports diverse environmental settings, including varying lighting conditions such as daytime, dusk, and nighttime, weather patterns such as clear skies, overcast, and light snow, as well as seasonal changes, allowing drone agents to train in environments that closely resemble real-world conditions. Equipped with built-in front, rear, left, right, and top multi-view cameras, the platform can generate high-resolution data such as RGB images, Depth images, and target segmentation maps, providing rich visual input for scene understanding and spatial modeling. Additionally, the AerialVLN Simulator supports dynamic flight operations for drones, offering precise control over their 3D position, orientation, and speed, while allowing the execution of complex maneuvers such as turns, climbs, and obstacle avoidance, ensuring smooth and flexible flight actions. Based on the ``Real-to-Sim-to-Real" design philosophy, the platform significantly narrows the gap between virtual environments and real-world applications, making it especially suitable for research and optimization in core drone tasks such as scene perception, spatial reasoning, path planning, and motion decision-making.

\subsubsection{Embodied City}

Embodied City \cite{gao2024embodied} is an advanced high-fidelity 3D urban simulation platform specifically designed for the evaluation and development of embodied intelligence. Its core feature is the realistic virtual environments built based on real-world urban areas, such as commercial districts in Beijing, including highly detailed building models, street networks, and dynamic simulations of pedestrian and vehicular traffic. The platform uses Unreal Engine as its technical foundation, combining historical data with simulation algorithms to provide continuous perception and interaction capabilities for various embodied agents, such as drones and ground vehicles. By integrating the AirSim interface, it supports multimodal input and output, including RGB images, Depth images, LiDAR, GPS, and IMU data, facilitating motion control and environmental exploration within simulations. The design encompasses five task areas: scene understanding, question answering, dialogue, visual language navigation, and task planning. Through an easy-to-use Python SDK and an online platform, users can conveniently remotely access and test a variety of agent behaviors, while supporting real-time operation of up to eight agents.

\section{Advances of FM-based UAV systems}

Integrating AI algorithms such as machine learning and deep learning into UAV systems has become a mainstream trend. However, applying traditional AI models in UAV tasks still faces numerous challenges. First, these models typically rely on task-specific datasets for training, resulting in insufficient generalization capability and poor robustness when significant discrepancies exist between the actual scenarios and the training distributions. Additionally, traditional AI models are often optimized for single tasks, making them less effective in addressing the complex requirements of multi-task collaboration. Furthermore, these models exhibit notable limitations in human-machine interaction and task collaboration \cite{zhang2021understanding, crawshaw2020multi, gehrmann2019visual}.

The introduction of LLMs, VFMs, and VLMs inject novel intelligent capabilities into UAV systems through natural language understanding, zero-shot adaptation, multimodal collaboration, and intuitive human-machine interaction.

\begin{figure*}[ht]
 \centering
 \includegraphics[scale=0.12]{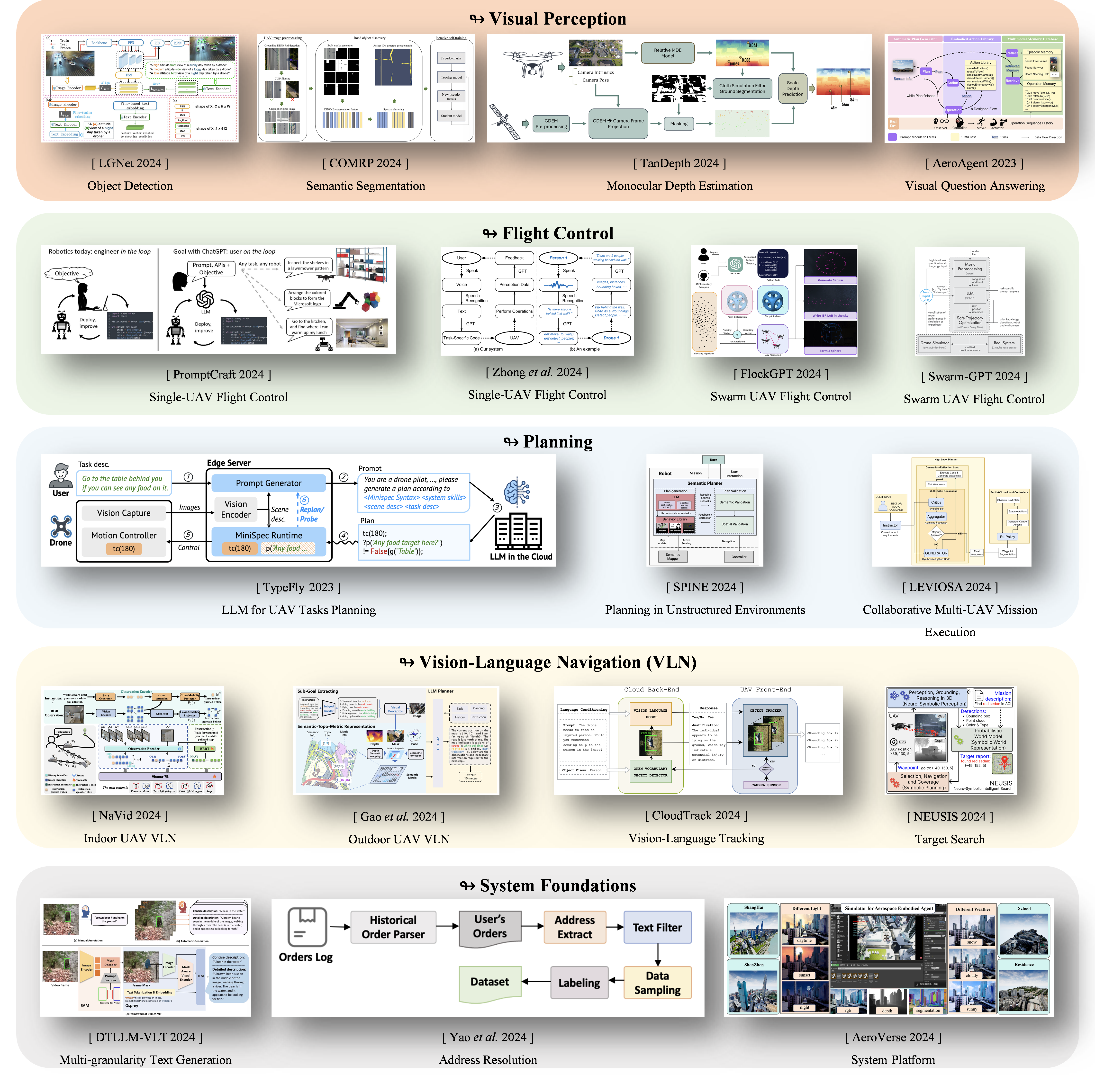}
 \caption{Typical works on FM-based UAV systems (Visual Perception: LGNet\cite{liu2024shooting}, CoMRP\cite{ma2024unsupervised}, TanDepth\cite{florea2024tandepth}, AeroAgent\cite{zhao2023agent}. Flight Control: PromptCraft\cite{vemprala2024chatgpt}, Zhong \emph{et al}.\cite{zhong2024safer}, FlockGPT\cite{lykov2024flockgpt}, Swarm-GPT\cite{jiao2023swarm}. Planning: TypeFlyv, SPINE\cite{ravichandran2024spine}, LEVIOSA\cite{aikins2024leviosa}. VLN: NaVid\cite{zhang2024navid}, Gao \emph{et al}.\cite{gao2024aerial}, CloudTrack\cite{blei2024cloudtrack}, NEUSIS\cite{cai2024neusis}, DTLLM-VLT\cite{li2024dtllm}, Yao \emph{et al}.\cite{yao2024can}, AeroVerse\cite{yao2024aeroverse}. ).}
 \label{fig:fm-uav}
\end{figure*}

This section explores existing research on integrating LLMs, VFMs, and VLMs into UAV systems and analyzes the advantages these technologies bring to different tasks. Several typical works are illustrated in Figure \ref{fig:fm-uav}. Based on the technical types and task characteristics, UAV-related tasks are categorized into the following types:
\begin{itemize}
    \item \textbf{Visual Perception:} These include object detection, semantic segmentation, depth estimation, visual caption, and VQA. Such tasks focus on environmental perception and semantic information extraction, serving as the foundation for high-level decision-making in UAV systems.
    \item \textbf{Vision-Language Navigation (VLN):} VLN represents a typical application of the deep integration of computer vision and natural language processing. Building on VLN tasks, more complex multimodal tasks, such as Vision-Language Tracking (VLT) and target search, have been developed. These tasks integrate multiple components, including perception, planning, decision-making, control, and human-machine interaction, forming the core framework for intelligent task execution in UAVs.
    \item \textbf{Planning:} This includes path optimization, task allocation, and adaptive task optimization in dynamic environments.
    \item \textbf{Flight Control:} These involve low-level control tasks such as attitude stabilization, path tracking, and obstacle avoidance.
    \item \textbf{Infrastructures:} This focuses on providing comprehensive technical and data support for UAV systems, including the development of integrated frameworks and platforms, as well as the creation and processing of high-quality datasets. These efforts not only enhance the efficiency of UAV applications in multimodal tasks but also provide critical support for foundational research and technological innovation in the UAV domain.
\end{itemize}

We provide a systematic comparison of relevant methods in Table~\ref{tab:methods_models} to offer a high-level overview of this rapidly evolving field. It should be noted that the ``Base Model’’ column in Table~\ref{tab:methods_models} lacks specific model names (e.g., GPT, LLM) in some cases, as the original references did not specify the exact model versions. For certain models, reference citations are included because detailed descriptions of the models were not provided in Section~\ref{sec:Preliminaries_on_Foundation_models}. Additionally, the notation ``LLMs or VLMs’’ indicates that multiple types of base models were tested in the corresponding method.

\begin{table*}[htbp]
    \centering
    \caption{Advances of FM-based UAV Systems in Various Tasks}
    \label{tab:methods_models}
    \setlength{\tabcolsep}{5pt} 
    \renewcommand{\arraystretch}{1} 
    \scriptsize 
    \newcolumntype{C}[1]{>{\centering\arraybackslash}m{#1}}
    \begin{tabular}{@{}C{2cm}C{2cm}C{4cm}C{2cm}C{6cm}@{}}
        \toprule
        \textbf{Category} & \textbf{Subcategory} & \textbf{Method / Model Name} & \textbf{Type} & \textbf{Base Model} \\
        \midrule
        \multirow{15}{*}{Visual Perception}
        & \multirow{6}{*}{Object Detection}
        & Li \emph{et al}.\cite{li2024benchmark}
        & VFM
        & CLIP \\
        &  & Ma \emph{et al}.\cite{ma2024unsupervised}
        & VFM
        & Grounding DINO + CLIP \\
        &  & Limberg \emph{et al}.\cite{limberg2024leveraging}
        & VFM+VLM
        & YOLO-World + GPT-4V \\
        &  & Kim \emph{et al}.\cite{kim2024weather}
        & VLM+VFM
        & LLaVA-1.5 + CLIP \\
        &  & LGNet\cite{liu2024shooting}
        & VFM
        & CLIP \\
        &  & \cite{sakaino2023dynamic}
        & VLM+VFM
        & BLIP-2 + OvSeg\cite{liang2023open} + ViLD\cite{gu2021open} \\
        \cmidrule(lr){2-5}
        & \multirow{2}{*}{Segmentation}
        & COMRP\cite{ma2024unsupervised}
        & VFM
        & Grounding DINO + CLIP + SAM + DINOv2 \\
        &  & CrossEarth\cite{gong2024crossearth}
        & VFM
        & DINOv2 \\
        \cmidrule(lr){2-5}
        & \multirow{1}{*}{Depth Estimation}
        & TanDepth\cite{florea2024tandepth}
        & VFM
        & Depth Anything \\
        \cmidrule(lr){2-5}
        & \multirow{6}{*}{Visual Caption/QA}
        & DroneGPT\cite{qiu2024dronegpt}
        & VLM+LLM+VFM
        & VISPROG + GPT-3.5 + Grounding DINO \\
        &  & de Zarzà \emph{et al} \cite{de2023socratic}.
        & LLM
        & BLIP-2 + GPT-3.5 \\
        &  & AeroAgent\cite{zhao2023agent}
        & VLM
        & GPT-4V \\
        &  & RS-LLaVA\cite{bazi2024rs}
        & VLM
        & LLaVA-1.5 \\
        &  & GeoRSCLIP\cite{zhang2024rs5m}
        & VFM
        & CLIP \\
        &  & SkyEyeGPT\cite{zhan2024skyeyegpt}
        & VFM+LLM
        & EVA-CLIP + LLaMA2 \\
        \midrule

        \multirow{13}{*}{VLN}
        & \multirow{2}{*}{Indoor}
        & NaVid\cite{zhang2024navid}
        & VFM+LLM
        & EVA-CLIP + Vicuna \\
        &  & VLN-MP\cite{hong2024only}
        & VFM
        & Grounding DINO / GLIP \\
        \cmidrule(lr){2-5}
        & \multirow{8}{*}{Outdoor}
        & Gao \emph{et al}.\cite{gao2024aerial}
        & VFM+LLM
        & Grounding DINO + TAP + GPT-4o \\
        &  & MGP\cite{lee2024citynav}
        & LLM+VFM
        & GPT-3.5 + Grounding DINO + MobileSAM \\
        &  & UAV Navigation LLM\cite{wang2024towards}
        & LLM+VFM
        & Vicuna + EVA-CLIP \\
        &  & GOMAA-Geo\cite{sarkar2024gomaa}
        & LLM+VFM
        & LLMs + CLIP \\
        &  & NavAgent\cite{liu2024navagent}
        & LLM+VFM+VLM
        & GLIP + BLIP-2 + GPT-4 + LLaMA2 \\
        &  & ASMA\cite{sanyal2024asma}
        & LLM+VFM
        & GPT-2 + CLIP \\
        &  & Zhang \emph{et al}.\cite{zhang2024demo}
        & VFM+LLM
        & GroundingDINO + LLM \\
        &  & Chen \emph{et al}.\cite{chen2023vision}
        & LLM
        & GPT-3.5 \\
        \cmidrule(lr){2-5}
        & \multirow{1}{*}{Tracking}
        & CloudTrack\cite{blei2024cloudtrack}
        & VFM+VLM
        & Grounding DINO + VLMs \\
        \cmidrule(lr){2-5}
        & \multirow{2}{*}{Target Search}
        & NEUSIS\cite{cai2024neusis}
        & VFM+VLM
        & HYDRA + CLIP + Grounding DINO + EfficientSAM \\
        &  & Say-REAPEx\cite{doschlsay}
        & LLM
        & GPT-4o-mini / Llama3 / Claude3 / Gemini \\
        \cmidrule(lr){1-5}
        
        \multirow{6}{*}{Planning}
        & \multirow{6}{*}{\_}
        & TypeFly\cite{chen2023typefly}
        & LLM
        & GPT-4 \\
        &  & SPINE\cite{ravichandran2024spine}
        & LLM+VFM+VLM
        & GPT-4 + Grounding DINO + LLaVA \\
        &  & LEVIOSA\cite{aikins2024leviosa}
        & LLM
        & Gemini 1.5 / GPT-4o \\
        &  & TPML\cite{cui2024tpml}
        & LLM
        & GPT / PengCheng Mind\cite{pengchengmind} \\
        &  & REAL\cite{tagliabue2023real}
        & LLM
        & GPT-4 \\
        &  & Liu \emph{et al}.\cite{liu2024multi}
        & LLM
        & GPT-4 \\
        \midrule

        \multirow{9}{*}{Flight Control}
        & \multirow{6}{*}{Single-agent}
        & PromptCraft\cite{vemprala2024chatgpt}
        & LLM
        & GPT \\
        &  & Zhong \emph{et al}.\cite{zhong2024safer}
        & LLM
        & GPT \\
        &  & Tazir \emph{et al}.\cite{tazir2023words}
        & LLM
        & GPT-3.5 \\
        &  & Phadke \emph{et al}.\cite{phadke2024integrating}
        & LLM
        & \_ \\
        &  & EAI-SIM\cite{liu2024eai}
        & LLM
        & GPT / PengCheng Mind\cite{pengchengmind} \\
        &  & TAIiST\cite{zhu2024taiist}
        & LLM
        & GPT-3.5 \\
        \cmidrule(lr){2-5}
        & \multirow{3}{*}{Swarm}
        & Swarm-GPT\cite{jiao2023swarm}
        & LLM
        & GPT-3.5 \\
        &  & FlockGPT\cite{lykov2024flockgpt}
        & LLM
        & GPT-4 \\
        &  & CLIPSwarm\cite{pueyo2024clipswarm}
        & VFM
        & CLIP \\
        \midrule
        
        \multirow{8}{*}{Infrastructures}
        & \multirow{8}{*}{\_}
        & DTLLM-VLT\cite{li2024dtllm}
        & VFM+LLM
        & SAM + Osprey + LLaMA / Vicuna \\
        &  & Yao \emph{et al}.\cite{yao2024can}
        & LLM
        & GPT-3.5 / ChatGLM \\
        &  & GPG2A\cite{arrabi2024cross}
        & LLM
        & Gemini \\
        &  & AeroVerse\cite{yao2024aeroverse}
        & VLM+LLM
        & VLMs + GPT-4 \\
        &  & Tang \emph{et al}.\cite{tang2024defining}
        & LLM
        & \_ \\
        &  & Xu \emph{et al}.\cite{xu2024emergency}
        & LLM
        & \_ \\
        &  & LLM-RS\cite{xiang2024real}
        & LLM
        & ChatGLM2 \\
        &  & Pineli \emph{et al}.\cite{pineli2024evaluating}
        & LLM
        & LLaMA3 \\
        \bottomrule
    \end{tabular}
\end{table*}

\subsection{Visual Perception}

\subsubsection{Object Detection}

In specific applications, Ma \emph{et al}. \cite{ma2024unsupervised} enhanced the accuracy of road scene detection in UAV imagery by integrating Grounding DINO \cite{liu2023grounding} and CLIP. Limberg \emph{et al}. \cite{limberg2024leveraging} utilized the combination of YOLO-World \cite{cheng2024yolo} and GPT-4V \cite{openai2024gpt4v} to achieve zero-shot human detection and action recognition in UAV imagery. Kim \emph{et al}. \cite{kim2024weather} employed LLaVA-1.5 \cite{liu2024improved} to generate weather descriptions for UAV images by combining visual features with language prompts such as weather and lighting conditions. Using a CLIP encoder, they fused image features with weather-related information. Based on this framework, weather-aware object queries were implemented, effectively leveraging weather information in object detection tasks, thereby significantly improving detection accuracy and robustness.

Notably, the multimodal representation capabilities of CLIP can generate high-quality domain-invariant features, providing strong support for training traditional object detection models. For example, LGNet \cite{liu2024shooting} introduced CLIP's multimodal features, significantly enhancing the robustness and performance of UAV object detection under diverse shooting conditions. Furthermore, LLMs, VLMs, and VFMs have accumulated extensive research experience in general object detection tasks, offering important insights for UAV object detection tasks. Examples include LLM-AR \cite{qu2024llms}, Han \emph{et al}. \cite{han2024few}, Lin \emph{et al}. \cite{lin2024generative}, ContextDET \cite{zang2024contextual}, and LLMI3D \cite{yang2024llmi3d}.

However, relying solely on VFMs or VLMs for object detection may lead to performance limitations in certain scenarios due to model hallucinations or inadequate task-specific adaptability \cite{huang2023survey, liu2024survey, favero2024multi}. While traditional deep learning models exhibit reliable performance in specific tasks, they lack cross-task generalization capabilities. A better solution is to adopt a ``large model + small model’’ collaborative architecture, leveraging the strong generalization capabilities of large models and the domain specialization of small models. For example, the Hidetomo Sakaino Visual Recognition Group \cite{sakaino2023dynamic} proposed a method combining DL models with VLMs for visibility and weather condition estimation. This method effectively addresses challenges such as scale variation, perspective changes, and environmental disturbances in image processing, including sky background interference and long-distance small object detection. It demonstrated outstanding robustness and stability across various environments and weather conditions.

\subsubsection{Semantic Segmentation}

The introduction of VLMs and VFMs into UAV semantic segmentation tasks has infused the field with new technological momentum. These models can efficiently perform zero-shot semantic segmentation while flexibly defining and guiding segmentation tasks through natural language interactions, demonstrating exceptional potential to meet diverse scenario requirements. For example, COMRP \cite{ma2024unsupervised} focuses on parsing road scenes in high-resolution UAV imagery. Its method first utilizes Grounding DINO \cite{liu2023grounding} and CLIP \cite{radford2021learning} to extract road-related regions and automatically generates segmentation masks using SAM \cite{kirillov2023segment}. Then, features are extracted using ResNet \cite{he2016deep} and DINOv2 \cite{oquab2023dinov2}, and mask feature vectors are clustered using a spectral clustering method to generate pseudo-labels. These labels are used to train a teacher model, iteratively optimizing the performance of a student model. COMRP eliminates the dependence on manual annotations, providing an efficient and automated solution for UAV road scene parsing. Additionally, CrossEarth \cite{gong2024crossearth} is a cross-domain generalization semantic segmentation VFM designed for the remote sensing field. It combines two complementary strategies: Earth-style injection and multi-task training, significantly enhancing cross-domain generalization capabilities. Earth-style injection incorporates diverse styles from the Earth domain into the source domain data, extending the distribution range of the training data. Multi-task training leverages a shared DINOv2 backbone to optimize both semantic segmentation and mask image modeling tasks simultaneously, enabling the learning of robust semantic features.

\subsubsection{Depth Estimation}

One of the core functionalities of UAV perception systems is to perform 3D modeling of terrain and natural environments, thereby generating consistent and accurate 3D geometric representations of the flight area.  In this context, MDE has emerged as a promising solution due to its advantages in both efficiency and accuracy \cite{florea2022survey, chang2023self, yu2023scene}. Florea \emph{et al}. \cite{florea2024tandepth} proposed the TanDepth framework, which combines the relative depth estimation of the Depth Anything\cite{yang2024depth2} model with Global Digital Elevation Model (GDEM) data to generate high-precision Depth images with real-world dimensions using a scale recovery method. Experimental results on multiple UAV datasets demonstrate that TanDepth exhibits outstanding accuracy and robustness in complex terrains and dynamic flight environments. This approach opens up new technological directions for UAV depth estimation tasks, particularly showcasing its efficiency and adaptability in scenarios lacking high-precision depth sensors.

\subsubsection{Visual Caption and VQA}

Traditional visual caption and VQA methods often separate visual feature extraction and language generation, which can be limiting in complex scenarios and for fine-grained descriptions, due to model constraints and misaligned multimodal features \cite{zhou2020unified, hu2022scaling}. However, with the rise of VLMs and VFMs, these models jointly learn vision and language representations, improving the understanding of complex cross-modal data. Pretrained on large multimodal datasets, VLMs and VFMs excel in generalizing across tasks and generating detailed descriptions in complex scenarios, making them highly adaptable to open-domain tasks \cite{liu2024visual, alayrac2022flamingo, li2022blip, li2023blip2, maaz2023video, gupta2023visual}.

In UAV visual captioning and VQA tasks, research focuses on two main directions: the first involves leveraging existing VLMs and VFMs in a zero-shot manner to adapt them to specific UAV scenarios; the second involves fine-tuning VLMs and VFMs with domain-specific data to create specialized models for UAV applications. Both directions aim to enhance UAVs' visual perception, semantic reasoning, and task execution capabilities in complex environments, contributing to more intelligent and user-friendly human-machine interaction.

For the first research direction, several studies have explored combining existing VLMs and VFMs to adapt to UAV scenarios. For instance, Qiu \emph{et al}. \cite{qiu2024dronegpt} proposed the DroneGPT framework based on the visual reasoning model VISPROG \cite{gupta2023visual}, where GPT-3.5 \cite{brown2020language} converts user natural language queries into task logic codes. These codes invoke Grounding DINO \cite{liu2023grounding} to parse visual information and perform semantic reasoning, ultimately outputting clear and accurate visual question-answering results. De Zarzà \emph{et al}. \cite{de2023semantic} designed a framework combining BLIP-2 \cite{li2023blip2} with GPT-3.5 for efficient UAV video scene understanding and semantic reasoning, where BLIP-2 extracts preliminary semantic information from each video frame, and GPT-3.5 generates high-level scene descriptions. The AeroAgent architecture \cite{zhao2023agent} optimizes UAV visual question-answering modules from an agent perspective. Built on GPT-4V \cite{openai2024gpt4v}, it constructs a retrievable multimodal memory database (similar to the RAG framework), significantly improving comprehension and answer accuracy in complex scenarios while mitigating hallucination issues in generative models.

The second direction focuses on fine-tuning VLMs and VFMs with domain-specific data to optimize models for UAV remote sensing tasks, improving semantic understanding of remote sensing images. Traditional remote sensing methods rely heavily on domain expertise and manual annotation, limiting performance in complex scenarios. To overcome these challenges, Bazi \emph{et al}. \cite{bazi2024rs} fine-tuned the LLaVA-1.5 \cite{liu2024improved} model for remote sensing tasks, enabling subtitle generation and VQA for remote sensing images. Zhang \emph{et al}. \cite{zhang2024rs5m} introduced GeoRSCLIP, a model trained on the RS5M dataset, which showed strong performance in zero-shot classification, cross-modal retrieval, and semantic localization. SkyEyeGPT \cite{zhan2024skyeyegpt} is a unified framework that fine-tunes EVA-CLIP \cite{sun2023eva} and LLaMA2 \cite{touvron2023llama2} for remote sensing visual-language tasks, supporting applications like image description, VQA, and visual localization.

\subsection{VLN}

Recent advances in VLN have been largely driven by deep learning techniques, particularly with the use of VLMs and VFMs. These models leverage large-scale pretraining to learn aligned multimodal feature representations, which greatly enhance task understanding and performance, especially in dynamic and complex environments \cite{chu2025towards}. When applied to UAVs, VLN presents unique challenges and characteristics. UAV VLN involves path planning in 3D space which requires consideration of flight altitude and sophisticated 3D spatial reasoning. Additionally, UAV VLN tasks vary significantly depending on the environment: indoor environments often have more defined geometric constraints, simplifying mission planning, while outdoor environments introduce additional complexity due to the scale of open spaces and dynamic changes, making navigation more challenging.

\subsubsection{Indoor}

For indoor UAV VLN, Neuro-LIFT \cite{joshi2025neuro} utilizes the LLMs for interaction between humans and UAV planners. LLMs first judge the feasibility of the maneuvers from humans and then transform human language into high-level planning commands. NaVid \cite{zhang2024navid} utilizes EVA-CLIP \cite{sun2023eva} to extract visual features, combined with Q-Former \cite{li2022blip, li2023blip2} to generate visual tokens and geometric tokens. Cross-modal projection aligns visual and language features, while Vicuna-7B \cite{vicuna2023} interprets natural language instructions and generates specific navigation actions. The system relies solely on monocular video streams without requiring maps, odometry, or depth information. By encoding historical observations as spatiotemporal context, it enables real-time reasoning for low-level navigation actions, demonstrating exceptional path planning and dynamic adjustment capabilities in indoor environments. Moreover, multimodal prompting shows significant potential in UAV VLN tasks. Hong \emph{et al}. \cite{hong2024only} proposed the VLN-MP framework, which enhances task understanding through multimodal prompts, reduces ambiguities in natural language instructions, and supports diverse and high-quality prompt settings. This system generates landmark-related image prompts using a data generation pipeline, combined with Grounding DINO \cite{liu2023grounding} or GLIP \cite{li2022grounded}, while ControlNet \cite{zhang2023adding} enhances data diversity. Finally, the system fuses image and text features via a visual encoder and multi-layer Transformer modules to generate precise navigation actions.

\subsubsection{Outdoor}

For outdoor UAV VLN, Liu \emph{et al}. \cite{liu2023aerialvln} proposed AerialVLN, addressing the gap in aerial navigation research. This task requires UAVs to navigate to target locations based on natural language instructions and first-person visual perception, treating all unoccupied points as navigable regions without preconstructed navigation maps. Based on this task, Liu \emph{et al}. developed an extended baseline model built on conventional cross-modal alignment (CMA) navigation methods, providing an initial solution for aerial navigation. Subsequent research incorporated LLMs to enhance task performance. For example, Gao \emph{et al}. \cite{gao2024aerial} designed an LLM-based end-to-end UAV VLN framework. This system uses GPT-4o to decompose natural language instructions into multiple sub-goals and combines Grounding DINO \cite{liu2023grounding} and Tokenize Anything (TAP)\cite{pan2025tokenize} to extract semantic masks and visual information. RGB images and Depth images are transformed into a semantic-topological-metric representation (STMR). With designed multimodal prompts, including task descriptions, historical trajectories, and semantic matrices, GPT-4o performs chain-of-thought reasoning to generate navigation actions (direction, rotation angle, and movement distance), significantly improving navigation success rates on the AerialVLN dataset. OpenFLY \cite{Gao2025OpenFlyAV} integrates several typical simulators to automate the data generation process and provides 15.6K vocabulary and 100K trajectories. 

Other notable studies include the CityNav dataset and its accompanying model MGP proposed by Lee \emph{et al}. \cite{lee2024citynav}. MGP uses GPT-3.5 \cite{brown2020language} to interpret landmark names, spatial relationships, and task goals, combining Grounding DINO \cite{liu2023grounding} and MobileSAM \cite{zhang2023faster} to generate high-precision target regions for navigation map construction and path planning. Wang \emph{et al}. \cite{wang2024towards} developed a system framework for UAV VLN, introducing the novel benchmark task UAV-Need-Help and constructing a related dataset via the OpenUAV simulation platform. Their UAV Navigation LLM, based on Vicuna-7B \cite{vicuna2023} and EVA-CLIP \cite{sun2023eva}, extracts visual features and employs a hierarchical trajectory generation mechanism for efficient natural language navigation. GOMAA-Geo\cite{sarkar2024gomaa} framework focuses on multimodal active geolocalization tasks by integrating various LLMs with CLIP\cite{radford2021learning}. It fully leverages multimodal target descriptions (such as natural language, ground images, and aerial images) and visual cues to achieve efficient and accurate target localization, demonstrating excellent zero-shot generalization capabilities. The NavAgent\cite{liu2024navagent} framework incorporates advanced models such as LLaMA2\cite{touvron2023llama2}, BLIP-2\cite{li2023blip2}, GPT-4\cite{achiam2023gpt}, and GLIP\cite{li2022grounded}. Parsing natural language navigation instructions to extract landmark descriptions and utilizing a fine-tuned landmark recognition module achieve precise landmark localization in panoramic images. This framework excels in path planning and navigation tasks in urban outdoor scenarios, providing robust technical support for UAV navigation in complex environments. Related studies, such as ASMA \cite{sanyal2024asma}, Zhang \emph{et al}. \cite{zhang2024demo}, and Chen \emph{et al}. \cite{chen2023vision} also explore UAV VLN solutions for outdoor environments and are worth further attention.


\subsubsection{VLT}

The VLT task aims to achieve continuous target tracking based on multimodal inputs while dynamically adjusting flight paths to address challenges such as target occlusion and environmental interference. Li \emph{et al}. \cite{li2024benchmark} introduced the UAVNLT dataset and developed a baseline method for UAV natural language tracking (TNL). The visual localization module in this method employs CLIP \cite{radford2021learning}, leveraging its multimodal features to precisely locate the target in the first frame. Similar to VLN tasks, VLT tasks integrate natural language descriptions with target bounding boxes, using natural language as auxiliary information to reduce ambiguities introduced by bounding boxes. The natural language descriptions in the TNL system clearly specify target attributes, helping the system accurately identify and track targets in complex scenarios, thereby effectively addressing tracking challenges in dynamic environments. Blei \emph{et al}. \cite{blei2024cloudtrack} proposed CloudTrack, an open-vocabulary target detection and tracking system for UAV search and rescue missions. This system adopts a cloud-edge collaborative architecture, combining Grounding DINO \cite{liu2023grounding} with VLMs to parse semantic descriptions, enabling the detection and filtering of complex targets. CloudTrack provides reliable technical support for intelligent UAV perception and dynamic task execution in resource-constrained environments, showcasing the potential of multimodal technologies in UAV intelligent missions.

\subsubsection{Target Search}

The target search task integrates multimodal target perception and intelligent mission planning, representing a complex high-level autonomous UAV mission. It can be viewed as a combination of ``VLN + Object Detection + Efficient Path Planning.’’ Compared to traditional VLN tasks, target search requires UAVs to efficiently perceive and locate targets while navigating \cite{wu2019uav, hou2023uav}.

Cai \emph{et al.} \cite{cai2024neusis} proposed the NEUSIS framework, a neural-symbolic approach for target search tasks in complex environments, enabling UAVs to perform autonomous perception, reasoning, and planning under uncertainty. The framework comprises three main modules: First, the Perception, Localization, and 3D Reasoning Module (GRiD) integrates VFMs and neural-symbolic methods, such as HYDRA \cite{ke2024hydra} for dynamic visual reasoning, CLIP \cite{radford2021learning} for target attribute classification, Grounding DINO \cite{liu2023grounding} for open-set target localization, and EfficientSAM \cite{xiong2024efficientsam} for efficient instance segmentation, to accomplish tasks like target detection, attribute recognition, and 3D projection. Second, the Probabilistic World Model Module employs Bayesian filtering and distribution ranking mechanisms to maintain probabilistic target maps and 3D environmental representations by fusing noisy data, thus supporting dynamic target localization and reliable report generation. Finally, the Selection, Navigation, and Coverage Module (SNaC) utilizes high-level region selection, mid-level path navigation, and low-level area coverage. Through the A$^{*}$ algorithm and belief map-based optimization methods, it generates efficient path planning schemes, ensuring the UAV maximizes target search tasks within limited time constraints. Döschl \emph{et al}. \cite{doschlsay} introduced the Say-REAPEx framework for online mission planning and execution in UAV search-and-rescue tasks. This framework uses GPT-4o-mini as the primary language model and tests Llama3 \cite{dubey2024llama}, Claude3 \cite{claude3modelcard2024}, and Gemini \cite{reid2024gemini} for parsing natural language mission instructions. It dynamically updates mission states using observational data and generates corresponding action plans. The framework also employs online heuristic search to optimize UAV mission paths, significantly enhancing real-time responsiveness and autonomous decision-making in dynamic environments. Say-REAPEx provides efficient and reliable technical solutions for complex tasks.

\subsection{Planning}

Traditional UAV mission planning algorithms face significant challenges in adaptability and coordination in complex dynamic environments. Task planning for multi-UAV systems must comprehensively consider the capabilities, limitations, and sensing modes of each UAV while satisfying constraints such as energy consumption and collision avoidance to achieve efficient mission allocation and path planning \cite{bethke2008uav, zhou2018mobile}. However, despite the new technical approaches provided by deep learning, these methods still exhibit limitations, such as heavy reliance on large-scale annotated data, insufficient real-time adaptation to environmental dynamics, and limited capability to handle unexpected situations or undefined fault modes. Additionally, models trained for fixed missions or environments often struggle to generalize well to different scenarios \cite{chang2020coactive, mao2024dl, yang2019application}.

LLMs, leveraging the CoT framework \cite{wei2022chain}, can decompose complex missions into a series of clear and executable subtasks, thereby providing a well-defined planning path and logical framework. With the advantages of in-context learning and few-shot learning, LLMs can flexibly adapt to diverse mission requirements and rapidly generate efficient planning strategies even without large-scale annotated data \cite{shen2024hugginggpt, khot2022decomposed}. Furthermore, LLMs’ outstanding performance in natural language understanding and generation enables real-time collaboration with operators through language instructions, significantly enhancing the intelligence and operational flexibility of mission planning.

AutoHMA-LLM \cite{yang2025autohma} proposes a cloud-edge framework with LLMs (Llama2 or GPT-4) for task coordination and collaborative execution of heterogeneous agents such as UAVs, robots, and intelligent cars. ACMA \cite{han2025agent} designs a multi-agent system with LLMs to coordinate UAV-based HAPS (High Altitude Platform Station) and ensure communication quality under occasional events. TypeFly \cite{chen2023typefly} uses GPT-4 \cite{achiam2023gpt} to parse natural language instructions provided by users and generate precise mission planning scripts. It also introduces a lightweight mission planning language (MiniSpec) to optimize the number of tokens required for mission generation, thereby improving mission generation efficiency and response speed. The framework integrates a visual encoding module for real-time environmental perception and dynamic mission adjustment and includes a ``Replan’’ mechanism to handle environmental changes during execution. SPINE \cite{ravichandran2024spine}, designed for mission planning in unstructured environments, combines GPT-4 and semantic topological maps to reason and dynamically plan from incomplete natural language mission descriptions. The framework employs Grounding DINO \cite{liu2023grounding} for object detection, LLaVA \cite{liu2024visual, liu2024improved} to enrich semantic information, and uses a Receding Horizon Framework to decompose complex missions into executable paths, enabling dynamic adjustments and efficient execution. LEVIOSA \cite{aikins2024leviosa} generates UAV trajectories through natural language, using Gemini \cite{team2023gemini, reid2024gemini} or GPT-4o to parse user text or voice inputs, translating mission requirements into high-level waypoint planning. The framework combines reinforcement learning with a multi-critic consensus mechanism to optimize trajectories, ensuring that the plans meet safety and energy efficiency requirements. It achieves end-to-end automation from natural language to 3D UAV trajectories, supporting dynamic environment adaptation and collaborative multi-UAV mission execution. UAV-VLA \cite{sautenkov2025uav} uses LLM to generate plans and actions for aerial tasks with the aid of VLMs processing satellite images. Similar studies include TPML \cite{cui2024tpml}, REAL \cite{tagliabue2023real}, and the work by Liu \emph{et al}. \cite{liu2024multi}, which further expands the applications of LLMs in UAV mission planning.

\subsection{Flight Control}

UAV flight control tasks are generally categorized into two types: single-UAV flight control and swarm UAV flight control. In single-UAV flight control, imitation learning and reinforcement learning methods have gradually become mainstream, demonstrating significant potential in enhancing the intelligence of control strategies \cite{tejaswi2022constrained, shukla2020imitation, choi2020imitation}. However, these methods typically rely on large-scale annotated data and face limitations in real-time performance and safety. In swarm UAV flight control, techniques such as multi-agent reinforcement learning and Graph Neural Networks (GNNs) provide powerful modeling capabilities for multi-UAV collaborative tasks, showing advantages in scenarios such as formation flying, task allocation, and dynamic obstacle avoidance \cite{wang2024enhancing, du2024distributed}. Nevertheless, these approaches still encounter significant challenges in communication delays, computational complexity, and global optimization capabilities.

Compared to traditional methods, LLM-based flight control introduces entirely new possibilities to the field. Leveraging few-shot learning capabilities, LLMs can quickly adapt to new task requirements; their in-context learning abilities enable models to dynamically analyze task environments and generate high-level flight strategies. Furthermore, semantic-based natural language interaction significantly enhances human-machine collaboration efficiency, supporting mission planning, real-time decision-making, and complex environment adaptation in UAVs. Although this research direction is still in its early exploratory stage, it has already shown tremendous potential in task scenarios requiring semantic understanding and high-level decision-making.

In the domain of single-UAV flight control, early studies laid an important foundation for applying LLMs to this task. For example, Courbon \emph{et al}. \cite{courbon2010vision} proposed a vision-based navigation strategy that uses a monocular camera to observe natural landmarks, building a visual memory and enabling autonomous navigation in unknown environments by matching current visual images with pre-recorded keyframes. Vemprala \emph{et al}. \cite{vemprala2024chatgpt} developed the PromptCraft platform, a pioneering work applying LLMs to UAV flight control. This platform integrates ChatGPT with the Microsoft AirSim \cite{shah2018airsim} simulation environment. By designing flight control-specific prompts and combining the ChatGPT API with the AirSim API, it enables natural language-driven flight control. Prompt design plays a critical role in this process, directly impacting the accuracy of task understanding and instruction generation. Similar studies include explorations by Zhong \emph{et al}. \cite{zhong2024safer}, Tazir \emph{et al}. \cite{tazir2023words}, and Phadke \emph{et al}.\cite{phadke2024integrating}, as well as the development of frameworks like EAI-SIM \cite{liu2024eai} and TAIiST \cite{zhu2024taiist}.

In the domain of swarm UAV flight control, Jiao \emph{et al}. \cite{jiao2023swarm} proposed the Swarm-GPT system, which combines LLMs with model-based safe motion planning to build an innovative framework for swarm UAV flight control. This system uses GPT-3.5 \cite{brown2020language} to generate time-series waypoints for UAVs and optimizes the paths through a safety planning module to satisfy physical constraints and collision avoidance requirements. Swarm-GPT allows users to dynamically modify flight paths through re-prompting, enabling flexible formation and dynamic adjustment of UAV swarms. Additionally, the system demonstrated the safety of trajectory planning and the artistic effects of formation performances in simulation environments. Similar research includes FlockGPT \cite{lykov2024flockgpt} and CLIPSwarm \cite{pueyo2024clipswarm}, which explore automated and creative control schemes to enhance the efficiency and operability of UAV swarm performances.

\subsection{Infrastructures}

The construction and processing of datasets are particularly critical in the foundational research of UAV systems. High-quality data resources and well-established data processing workflows are essential to ensuring the efficient application of LLM, VLM, and VFM technologies in UAV tasks. These research efforts not only lay a solid foundation for the application of UAVs in multimodal tasks but also provide strong support for technological innovation and methodological advancements in related fields.

DTLLM-VLT \cite{li2024dtllm} is a framework designed to enhance VLT performance through multi-granularity text generation. The framework uses SAM \cite{kirillov2023segment} to extract segmentation masks of targets, combined with Osprey \cite{yuan2024osprey} to generate initial visual descriptions. LLaMA \cite{touvron2023llama, touvron2023llama2, dubey2024llama} or Vicuna \cite{chiang2023vicuna} then generates four types of granular text annotations: initial brief descriptions, initial detailed descriptions, dense brief descriptions, and dense detailed descriptions, covering target categories, colors, actions, and dynamic changes. These high-quality text data significantly enhance semantic support for multimodal tasks, improving tracking accuracy and robustness while reducing the time and cost of semantic annotation. Yao \emph{et al}. \cite{yao2024can} developed the CNER-UAV dataset for fine-grained Chinese Named Entity Recognition in UAV delivery systems, leveraging GPT-3.5 \cite{brown2020language} and ChatGLM \cite{du2021glm, zeng2022glm, glm2024chatglm} to achieve precise address information recognition.

A noteworthy challenge in UAV systems is the high cost and labor-intensive effort of acquiring aerial imagery. To address this, Arrabi \emph{et al}. \cite{arrabi2024cross} proposed the GPG2A model, which synthesizes aerial imagery from ground images using Ground-to-Aerial (G2A) techniques, overcoming the generation challenges posed by significant viewpoint differences. The model employs a two-stage generation framework: the first stage uses ConvNeXt-B \cite{liu2022convnet} to extract ground image features and applies the polar coordinate transformation to generate bird's-eye view (BEV) layout maps for capturing scene geometry explicitly. The second stage introduces a diffusion model to generate high-quality aerial imagery by combining BEV layout maps with textual descriptions. The textual descriptions are generated by Gemini \cite{team2023gemini, reid2024gemini} and optimized into Dynamic Text Prompts using BERT \cite{devlin2018bert}, enhancing the semantic relevance and scene consistency of the generated imagery. This approach effectively addresses the challenges of viewpoint transformation and provides an innovative solution for efficiently acquiring aerial imagery, offering significant practical value.

In terms of frameworks and platforms, related research demonstrates diverse development directions. Yao \emph{et al}. \cite{yao2024aeroverse} proposed AeroVerse, a highly referential platform designed as an aviation intelligence benchmark suite for UAV agents. AeroVerse integrates simulators, datasets, task definitions, and evaluation methodologies to advance UAV technologies in perception, cognition, planning, and decision-making. Its system architecture includes a high-precision simulation platform, AeroSimulator, based on Unreal Engine and AirSim. AeroSimulator generates multimodal datasets spanning real and virtual scenes and provides fine-tuned datasets customized for five core tasks: scene perception, spatial reasoning, navigation exploration, mission planning, and motion decision-making.

Additionally, several innovative frameworks combine LLMs with UAV-specific tasks. For example, Tang \emph{et al}. \cite{tang2024defining} developed a safety assessment framework for UAV control; Xu \emph{et al}. \cite{xu2024emergency} designed an emergency communication network optimization framework for UAV deployment in dynamic environments; LLM-RS \cite{xiang2024real} focuses on UAV air combat simulation tasks, incorporating reward design and decision optimization to enhance system performance; Pineli \emph{et al}. \cite{pineli2024evaluating} proposed a UAV voice control framework, leveraging natural language processing technologies to maximize the potential of human-machine interaction. These works contribute to the development of UAV technologies from various dimensions, forming essential support for UAV intelligence and task diversification.

\section{Application scenarios of FMs-based UAVs}

\begin{figure*}[ht]
 \centering
 \includegraphics[scale=2.1]{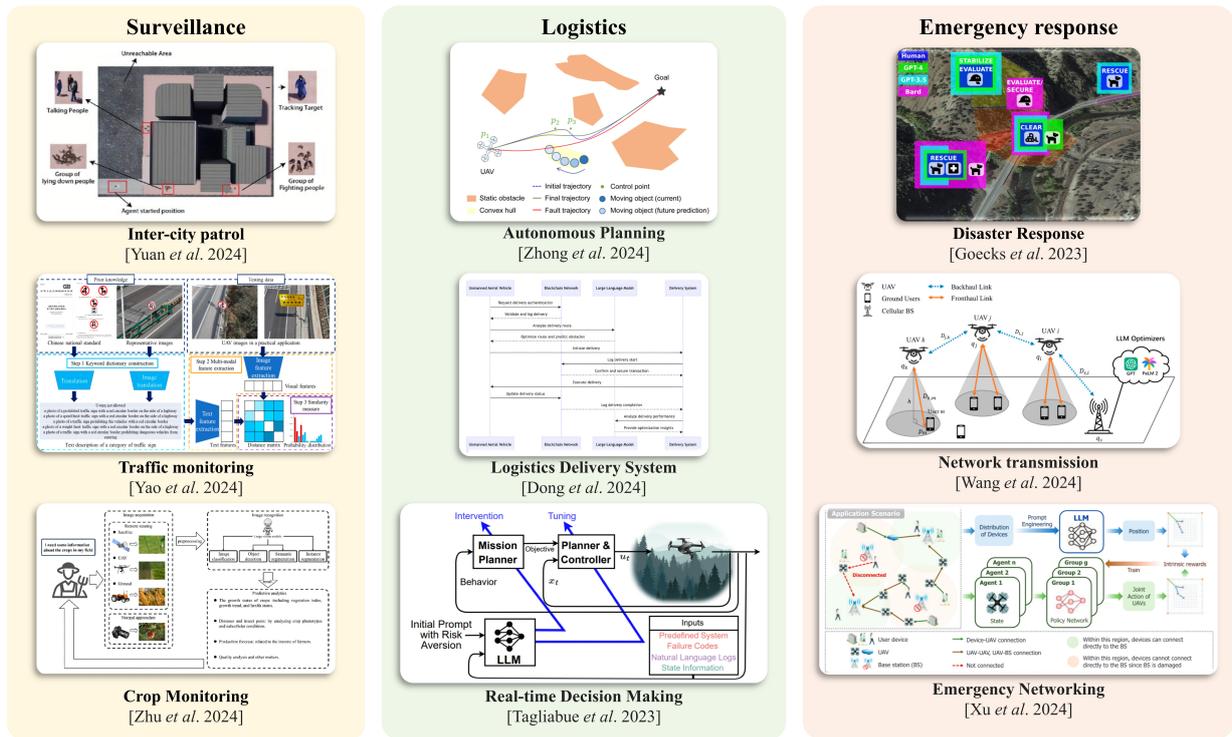}
 \caption{Typical applications on the integration of UAVs and FMs. (Surveillance: Yuan \emph{et al}. \cite{yuan2024patrol}, Yao \emph{et al}. \cite{yao2024vision}, Zhu \emph{et al}. \cite{zhu2024harnessing}; Logistics: Zhong \emph{et al}. \cite{zhong2024safer}, Dong \emph{et al}. \cite{dong2024securing}, Tagliabue \emph{et al}. \cite{tagliabue2023real}; Emergency response: Goecks \emph{et al}. \cite{goecks2023disasterresponsegpt}, Wang \emph{et al}. \cite{wang2024multi}, Xu \emph{et al}. \cite{xu2024emergency})}
 \label{fig_application}
\end{figure*}

This section focuses on the practical application scenarios of combining UAVs with LLMs. LLMs provide advanced cognitive and analytical capabilities for multimodal data, including image, audio, text, and even video data. Compared to UAVs integrated with traditional machine learning algorithms, incorporating LLMs into UAV systems significantly enhances their environmental perception capabilities \cite{de2023semantic,wang2023uav}, enables smarter decision-making processes \cite{kuwertz2018applying}, and improves user experience by leveraging the strong comprehension abilities of LLMs in human-machine interaction \cite{phadke2024integrating,feng2023large}.

Based on existing literature, we introduce typical works on the integration of FMs with UAVs as illustrated in Figure \ref{fig_application}: Surveillance, Logistics and Emergency Response. These three categories presented are not exhaustive of all UAV applications but rather represent the current areas where the combination of UAV technology and advanced model capabilities has been particularly effective. They focus on improving three key capabilities: environmental perception, autonomous decision-making, and human-machine interaction.

\subsection{Surveillance}
For surveillance, UAVs are used for monitoring traffic scenarios, urban environments, and other regulatory tasks. Traditional methods for addressing UAV applications in monitoring tasks primarily rely on machine learning techniques. In recent years, substantial research has been conducted in this area, including vehicle trajectory monitoring \cite{mahajan2023treating}, road condition monitoring \cite{telikani2024machine,bisio2022systematic}, road-side units (RSUs) communication \cite{saputro2018drone}, and applications and management in urban scenarios \cite{dung2019developing}. However, Menouar \emph{et al}. \cite{menouar2017uav} pointed out that UAVs are expected to play a significant role in Intelligent Transportation Systems (ITS) and smart cities, but their effectiveness will depend on greater autonomy and automation. Similarly, Wang \emph{et al}. \cite{wang2023review} emphasized the importance of UAVs in urban management and highlighted challenges such as automation and human-machine interaction. The emergence of FMs has recently led to research exploring how the integration of FMs with UAVs can enhance their usability and task performance.

In urban scenario monitoring, Yao \emph{et al}. \cite{yao2024vision} deployed VLMs for monitoring the conditions of traffic signs using multimodal learning and large-scale pre-trained networks, achieving excellent results in both accuracy and cost efficiency. UAVs integrated with FMs excel in tasks such as vehicle detection, vehicle classification, pedestrian detection, cyclist detection, speed estimation, and vehicle counting. Yuan \emph{et al}. \cite{yuan2024patrol} proposed the ``Patrol Agent,'' which leverages VLMs for visual information acquisition and LLMs for analysis and decision-making. This enables UAVs to autonomously conduct urban patrolling, identification, and tracking tasks. Additionally, UAVs integrated with LLMs have demonstrated outstanding performance in other monitoring tasks. In the monitoring of agricultural crops, Zhu \emph{et al}. \cite{zhu2024harnessing} review the applications of LLMs and VLMs, concluding that these technologies can significantly assist farmers in enhancing productivity and yields.

\subsection{Logistics}
For logistics, UAVs enable intelligent processes throughout the entire logistics chain, from decision-making to route planning and final delivery \cite{tian2024Logvista}. The application of UAVs in logistics and delivery is a key area of current research. Jiang \emph{et al}. \cite{jiang2024optimisation} optimized UAV scheduling and route planning using advanced optimization algorithms. Huang \emph{et al}. \cite{huang2020scheduling} proposed a collaborative scheduling solution involving UAVs and public transportation systems, such as trams, which was proven to be NP-complete. They also introduced a precise algorithm based on dynamic programming to address this challenge. However, UAV logistics still faces several challenges. Wandelt \emph{et al}. \cite{wandelt2023aerial} identified two primary issues: autonomous navigation and human-machine interaction, as well as real-time data analysis. The introduction of FMs provides a novel approach to addressing these challenges, offering the potential to enhance UAVs’ real-time decision-making and planning capabilities through FMs’ reasoning and decision-making power. Additionally, FMs’ strong comprehension capabilities improve human-machine interaction, providing a better user experience.

For logistic applications with FMs, Tagliabue \emph{et al}. \cite{tagliabue2023real} proposed a framework called REAL, leveraging prior knowledge from LLMs and employing zero-shot prompting. This method significantly improved UAV adaptability and decision-making, improving positional control performance and real-time task decision-making. Luo \emph{et al}. \cite{luo2024language} utilized LLMs to process user-provided address information. As traditional methods struggle with fine-grained handling due to the lack of precision in user inputs, they fine-tuned LLMs to address this issue, thereby increasing the automation level and processing efficiency of UAV delivery systems. Zhong \emph{et al}. \cite{zhong2024safer} focused on autonomous UAV planning and proposed a vision-based planning system integrated with LLMs. Their system combines dynamic obstacle tracking and trajectory prediction to achieve efficient and reliable autonomous flight. Additionally, integrating LLMs enhanced human-machine interaction, improving the overall user experience. Dong \emph{et al}. \cite{dong2024securing} approached the problem from a supply chain perspective, presenting an innovative intelligent delivery system for UAV logistics. By incorporating blockchain technology, they ensured system security and transparency. Furthermore, they utilized LLMs for route optimization and dynamic task management, and provided customer support services through natural language interaction, offering a framework for developing secure and efficient UAV delivery systems in the future.

\subsection{Emergency Response}

UAVs possess inherent advantages in emergency response and disaster relief missions \cite{boroujeni2024comprehensive}. Their highly flexible operational capabilities make them suitable for most emergency scenarios. Jin \emph{et al}. \cite{jin2020research} analyzed the demands for UAV-based emergency response mechanisms, evaluating disaster types and the performance characteristics of UAVs, while providing recommendations. By equipping UAVs with different payloads and supplies, they can deliver customized support based on specific disaster scenarios and mission requirements. Goecks \emph{et al}. \cite{goecks2023disasterresponsegpt} introduced DisasterResponse GPT, an LLM-based model that leverages contextual learning to accelerate disaster response by generating actionable plans and rapidly updating and adjusting them in real-time, enabling fast decision-making. De Curtò \emph{et al}. \cite{de2023semantic} capitalized on UAVs’ ability to provide instantaneous visual feedback and high data throughput, developing a scene understanding model that combines LLMs and VLMs. This approach cost-effectively enhances UAVs’ real-time decision-making capabilities for handling complex and dynamic data. Furthermore, they integrated multiple sensors to autonomously execute complex tasks.

Beyond rescue missions, UAVs are increasingly studied as tools for establishing communication networks in response to connectivity challenges in disaster-stricken or remote areas. Such networks support network-dependent tasks and offline emergency responses. Fourati \emph{et al}. \cite{fourati2021artificial} highlighted the critical role of AI in communication engineering, including applications such as flow prediction and channel modeling. Xu \emph{et al}. \cite{xu2024emergency} utilized UAVs as mobile access points to assist urban communication systems with emergency network deployment in disaster scenarios. They further employed LLMs to enhance the modeling process and accelerate optimization workflows. Wang \emph{et al}. \cite{wang2024multi} optimized UAV swarm deployment using structured prompts with LLMs. Compared to traditional methods, their approach reduces the number of iterations while ensuring strong network connectivity and quality of service through precise UAV positioning. The LLM-driven framework simplifies operational challenges for UAV network operators, paving the way for their application in more complex real-world scenarios.

\section{Agentic UAV: The General Pipeline Integrating FMs with UAV Systems}
\begin{figure*}[!ht]
 \centering
 \includegraphics[width=1\textwidth, trim=0cm 10cm 0cm 0cm, clip]{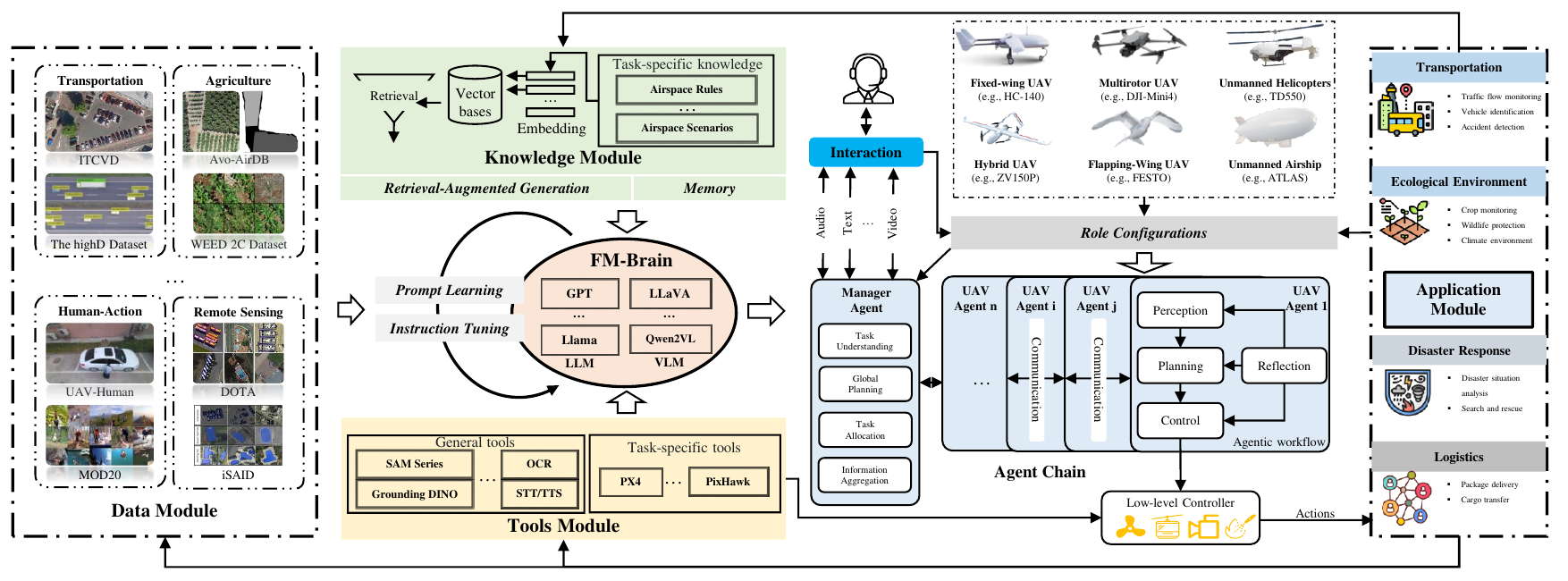}
 \caption{The framework of Agentic UAV.}
 \label{fig_agenticuav}
\end{figure*}

This section systematically explores the integration of LLMs and VLMs into traditional UAV pipelines and tasks. From the perspective of AI agents, we propose the framework of Agentic UAV that combines FMs with UAV systems, as illustrated in Figure \ref{fig_agenticuav}. The framework comprises five key components: data module, knowledge module, tools module, FM module, and agent module. The data module focuses on creating new datasets or adapting existing data to formats suitable for fine-tuning and training FMs tailored to UAV-specific tasks. The knowledge module stores domain-specific information, such as airspace regulations and scenario libraries, essential for UAV operations. The tools module includes domain-specific tools or APIs required to address UAV tasks, thereby extending the agent's problem-solving capabilities. The FM module concentrates on fine-tuning FMs to enhance their adaptation and performance in UAV-related domains. The agent module is designed to create workflows incorporating perception, planning, and action for UAV tasks. This module also establishes reflective mechanisms to optimize processes based on feedback from task execution. Additionally, considering the frequent use of UAV swarm, the agent module integrates multi-agent designs, interaction, and communication units. To coordinate and manage these agents, the framework introduces a manager agent responsible for global task planning and allocation. Each of these modules is elaborated upon in the following subsections.

In contrast to traditional UAV systems, which primarily operate based on predefined, rule-based processes, Agentic UAVs possess the ability to continuously learn and adapt. The agent module allows the UAV to perceive its environment, plan its actions based on real-time data, and execute tasks independently. This dynamic decision-making process enables UAVs to respond more effectively to unforeseen challenges, such as sudden obstacles or changes in mission parameters. In addition, Agentic UAVs provide human-like task understanding and interaction capabilities, offering users a more convenient and intuitive service experience.

\subsection{Data Module}

The data module is designed to convert UAV-related data into formats such as captions, question-answering, or chain-of-thought, making it suitable for fine-tuning and training FMs tailored to UAV-specific tasks. For instance, benchmark datasets tailored for UAV navigation and geolocation tasks have been developed by Chu \emph{et al.} \cite{chu2025towards}, which extends existing resources with text-image-bounding box annotations to improve geolocation accuracy. Similarly, Yao \emph{et al.} \cite{yao2024can} introduced a fine-grained Chinese address recognition dataset for UAV delivery, enhancing navigation precision in urban contexts. Furthermore, in remote sensing applications, UAV imagery has been extensively utilized for tasks like object detection, semantic segmentation, and environmental monitoring, with multimodal large models significantly improving task efficiency and accuracy \cite{gong2024crossearth}.

The data processing flow for UAV-oriented LLMs and VLMs involves multiple stages. During the pre-training phase, data preparation focuses on building effective representations through image-text contrastive learning or generative language prediction to construct FM-ready formats. This process aims at injecting general and diverse UAV knowledge. In the fine-tuning phase, the focus shifts to domain-specific tasks, where single-modal or multi-modal data, such as descriptions or question-answer pairs, are constructed to transfer perceptual and decision-making capabilities to the model. Furthermore, to align UAV system outputs with human preferences, reinforcement learning fine-tuning datasets, including rejection sampling, can be utilized. For more complex tasks, such as planning and solving cluster flight missions, chain-of-thought datasets can be created to facilitate long-range reasoning tasks. This comprehensive approach ensures that UAV systems can learn and generalize effectively across various types of tasks.

\subsection{FM Module}

The FM module for UAV tasks focuses on two core aspects: selecting appropriate models and optimizing them for specific tasks. This modular approach ensures that UAV systems can handle diverse and complex scenarios effectively while maintaining efficiency in execution.

\subsubsection{Model Selection}  
The process begins with identifying the task type and determining whether the data involves single-modal or multimodal inputs. For language-based tasks, LLMs such as ChatGPT and LLama provide a robust foundation for reasoning, decision-making, and natural language interaction. For multimodal tasks, such as those involving visual and linguistic data, VLMs such as GPT-4V, LLaVa, and Qwen2-VL, are often ideal. These models serve as foundational components, providing a capability backbone for intelligent agents.

In addition to language and vision-based models, recent advancements have explored large 3D models, which are particularly relevant to UAVs operating in 3D environments. These models integrate FMs with capabilities for interpreting 3D data and planning tasks. For instance, Hong \emph{et al.}~\cite{hong20233d} proposed a 3D LLM capable of dense captioning, 3D question answering, and navigation using point clouds. Similarly, Agent3D-Zero~\cite{zhang2024agent3d} employs Set-of-Line Prompts (SoLP) to enhance scene geometry understanding by generating diverse observational perspectives. While most current research focuses on indoor and closed environments, expanding these models to open and dynamic UAV scenarios presents exciting future opportunities.

\subsubsection{Model Optimization}
Once the base model is selected, it is then optimized and adapted to meet UAV-specific requirements through various prompting and fine-tuning techniques. Prompt engineering serves as an effective method by designing task-specific templates that embed mission background knowledge, such as objectives, environmental features, and task decomposition, into the model’s interactions. This approach ensures that the model is primed for UAV-related tasks. Few-shot learning can complement prompt engineering by providing carefully curated examples, enabling the model to better understand task-specific goals. For more complex UAV challenges, prompt learning through the TT approach \cite{wei2022chain} offers a powerful tool. CoT decomposes tasks into sequential subtasks, enhancing the model’s reasoning and execution capabilities.

For the knowledge that is absent in the pre-training stage, fine-tuning techniques become essential for optimizing the model. Instruction fine-tuning adapts the model by generating domain-specific datasets tailored to UAV tasks. Techniques like LoRA \cite{hu2021lora} can optimize the model by fine-tuning only a subset of parameters, thereby maintaining computational efficiency while improving task-specific performance. Additionally, layer-freezing techniques help preserve pre-trained knowledge and prevent overfitting, especially when working with smaller, task-specific datasets. 
To align the model's behavior with human preferences and operational requirements, Reinforcement Learning from Human Feedback (RLHF) \cite{casper2023open} can be employed. RLHF incorporates reward signals based on human feedback, guiding the model to adapt dynamically to challenges involving human values. For complex task, Reinforcement Fine Tuning (RFT) is crucial for constructing robust long-range reasoning capabilities. For example, during the fine-tuning phase, the model is trained to generate reasoning chains, helping it break down complex UAV tasks.

\subsection{Knowledge Module}

Retrieval-augmented generation (RAG) is an emerging technology that integrates retrieval and generation capabilities. Its core functionality lies in retrieving relevant information from a knowledge base and fusing it with the output of a generative model, thereby enhancing the accuracy and domain adaptability of generated results. RAG models leverage a retrieval module to obtain information pertinent to the input content from external knowledge repositories and incorporate it as context for the generative module. This approach improves the quality and reliability of generated outputs. Unlike traditional generative models, RAG introduces a real-time retrieval mechanism to mitigate the ``hallucination" problem, wherein a model generates incorrect or fabricated information due to insufficient background knowledge. Moreover, the modular architecture of RAG allows for independent updates of the knowledge base and generative model, increasing system flexibility and ensuring the timeliness and accuracy of the information used in generation. Consequently, RAG demonstrates significant potential in tasks requiring high specialization, real-time information processing, or personalized outputs.

Constructing RAG systems tailored for UAV-specific tasks is crucial because UAV operations involve diverse and complex scenarios. First, RAG can provide real-time access to up-to-date environmental data, such as meteorological conditions, terrain information, and air traffic updates, which are essential for tasks like flight planning and navigation. Second, integrating a domain-specific knowledge base into the RAG framework enables UAVs to perform advanced decision-making tasks, such as autonomous mission adjustments in dynamic environments or identifying unknown objects during surveillance missions. Finally, RAG can facilitate interaction with human operators by retrieving contextual data to clarify queries or enhance the interpretability of system decisions. For example, in UAV-based environmental monitoring tasks, RAG can retrieve historical data on pollution levels or land use patterns, combine this with current sensor data, and generate comprehensive reports. These capabilities illustrate how a well-constructed RAG framework can enhance the efficiency, accuracy, and adaptability of UAV systems, paving the way for more intelligent and autonomous UAV applications.

\subsection{Tools Module}

The Tools Module is designed to provide both general-purpose functionalities and task-specific capabilities to support UAV operations. 

\subsubsection{General Tools}  
General tools focus on broad, multimodal functionalities to enhance the UAV system's perception and interaction capabilities. Among these, VFMs serve as a cornerstone for addressing diverse visual tasks, leveraging their exceptional generalization and zero-shot learning capabilities. Unlike FMs that emphasize reasoning and decision-making, VFMs excel in understanding specific visual tasks, making them ideal as foundational tools rather than core ``FM-Brain" components.  

VFMs offer significant advantages in UAV missions by aligning with specific task requirements. For instance, the CLIP series is well-suited for object recognition and scene understanding tasks due to its robust multimodal alignment, enabling open-vocabulary object detection and classification. The SAM, renowned for its zero-shot segmentation capabilities, is ideal for image segmentation across varied environments and targets. Grounding DINO excels in object detection and localization tasks, providing efficient target tracking and detection in dynamic scenarios. These models can independently handle specific tasks or integrate with LLMs or VLMs to enhance UAV systems' intelligence in mission planning, navigation, and environmental perception.

Moreover, VFMs can be fine-tuned to adapt to UAV-specific scenarios. For instance, fine-tuning the Grounding DINO model on specialized datasets improves its performance in complex multi-target tracking tasks. Additionally, VFMs can collaborate with traditional machine learning or deep learning models to form a ``large model + small model" strategy, balancing generalization with task-specific efficiency. For example, VFMs extract global semantic information, while smaller models focus on fine-grained details, achieving an effective combination of global and local analyses.

Another innovative application of VFMs involves their use in generating instruction fine-tuning datasets for VLMs. By leveraging VFM outputs such as image captions, segmentation descriptions, and object depth information, these datasets can train VLMs for UAV-specific missions. For example, Chen \emph{et al.}~\cite{chen2024spatialvlm} created a 3D spatial instruction fine-tuning dataset using internet-scale spatial reasoning data from VFMs, training the SpatialVLM model. This approach highlights VFMs' potential to generate high-quality datasets for large models, significantly enhancing UAV systems' dynamic perception and mission planning capabilities.

\subsubsection{Task-Specific Tools}  
Task-specific tools are tailored to UAV-centric operations, focusing on flight control and mission execution. Key components include PX4 and Pixhawk, widely used open-source flight controllers. These tools provide UAVs with precise control, mission planning, and real-time adaptability, making them indispensable for complex aerial tasks. By combining these specialized tools with general functionalities, the UAV system achieves a high degree of flexibility and efficiency in addressing mission-specific challenges.

\subsection{Agent Module}
The Agent Module is designed to provide intelligent decision-making and task execution capabilities within the UAV system. It integrates both high-level coordination and task-specific agent workflows to optimize UAV operations in complex missions.

\subsubsection{Manager Agent}  
The Manage Agent is responsible for high-level task coordination and scheduling within the UAV swarm, ensuring that missions are executed efficiently across multiple UAVs. This agent takes on the role of global planning and overall task allocation, breaking down a large mission into smaller, manageable sub-tasks, which are then assigned to individual UAVs. Additionally, the Global Agent monitors the swarm's status and dynamically adjusts the distribution of tasks based on real-time feedback, ensuring that each UAV operates effectively within the context of the broader mission. 

\subsubsection{UAV-Specific Agentic Workflow}  
Each UAV in the swarm follows an autonomous Agentic Workflow that consists of a chain of agents designed to handle perception, planning, and control tasks. These agents operate in sequence, ensuring that each UAV processes the necessary data and executes its mission objectives effectively. The perception agent first processes sensor data, identifying obstacles, objects, and points of interest using advanced VFMs, such as CLIP for object recognition, SAM for segmentation, and Grounding DINO for localization. 

Next, the planning agent takes the data from the perception agent to generate optimized flight paths and task strategies, ensuring that the UAV can navigate the environment and complete the assigned mission efficiently. Finally, the control agent converts the plans into actionable commands, controlling the UAV’s flight and task execution. 

This workflow allows each UAV to operate independently while still contributing to the overall mission goals. Moreover, the UAV-Specific Agentic Workflow is adaptable to a wide variety of UAV missions, from search and rescue to surveillance, by fine-tuning the agents' capabilities according to the specific requirements of each task. This adaptability enhances the UAV's efficiency in handling complex, dynamic environments.

\subsubsection{Agent Collaboration and Adaptability}  
The collaboration between the Global Agent and UAV-specific agents is crucial for optimizing mission execution. The Global Agent provides high-level directives that guide the overall mission strategy. These directives are broken down into detailed execution plans by individual UAV agents, ensuring that each UAV can operate autonomously while contributing to the collective mission goal. The UAV agents communicate with the Global Agent to receive updated instructions and report progress, enabling continuous task adaptation and dynamic adjustments to the mission plan in response to real-time data and changing conditions.

Furthermore, UAV agents within the swarm can interact with each other to exchange information and coordinate their actions. This peer-to-peer communication enables the UAVs to adapt their behavior based on shared situational awareness, such as when multiple UAVs must avoid collisions or collaborate to accomplish a joint task. For example, one UAV might share its perception data with another to adjust flight paths or synchronize tasks in real-time. This interaction ensures that the UAV swarm operates cohesively, with each agent adjusting its actions based on both global guidance and local, real-time information from other agents.

\subsubsection{Discussion}

Despite the promising potential of Agentic UAVs, several challenges remain in their development and deployment. First, there is the issue of computational cost. The large parameter size and high computational demands of FMs require significant hardware resources, which can be a major limitation for real-time UAV operations. Furthermore, the response delay in processing complex tasks or generating answers can be problematic, especially for time-sensitive missions such as search and rescue, where immediate decision-making is crucial.

Second, security concerns pose another challenge. Agentic UAVs, like any AI-powered system, are prone to generating hallucinations (incorrect or fabricated information) especially when dealing with ambiguous or incomplete data. Such inaccuracies can lead to unsafe outcomes, which may be detrimental in critical applications like military operations or emergency response. Ensuring that the system can distinguish reliable from unreliable information, and implementing fail-safes to mitigate the risks of unsafe behavior, is crucial for the safe deployment of Agentic UAVs.

Third, there is a lack of foundational infrastructure to support large-scale deployment. The widespread use of Agentic UAVs requires reliable power sources, communication networks, and other essential infrastructure. For instance, continuous power supply and real-time communication are key to maintaining UAV operations, especially for long-duration or swarm-based tasks. Developing a robust infrastructure that can provide these capabilities at scale is essential for the practical deployment and sustainability of Agentic UAV systems in various industries.

\section{Conclusion}
This paper explores the promising integration of LLMs with UAVs, emphasizing the transformative potential of LLMs in enhancing UAV decision-making, perception, and reasoning capabilities. We begin by providing an overview of UAV system components and the underlying principles of large models, establishing the foundation for their integration. The paper then reviews the classification, research progress, and application scenarios of UAV systems enhanced by foundational LLMs. Additionally, we highlight key UAV-related datasets that support the development of intelligent UAV systems. Furthermore, we propose a forward-looking framework for the UAV field: Agentic UAVs, where multi-agent systems integrate knowledge and tool modules to create flexible UAVs capable of addressing complex tasks in dynamic environments. Looking ahead, quantitative acceleration techniques, such as model pruning and edge computing, are essential to reduce computational demands. Another crucial avenue of development is the creation of joint air, land, and sea unmanned systems that can operate cohesively in complex environments, enabling coordinated missions across multiple domains, such as disaster relief or military operations.

\section{Acknowledgement}
This work is partly supported by National Natural Science Foundation of China (62303460, 52441202), Beijing Natural Science Foundation-Fengtai Rail Transit Frontier Research Joint Fund (L231002), The Science and Technology Development Fund of Macau SAR (No. 0145/2023/RIA3 and 0093/2023/RIA2), and the Young Elite Scientists Sponsorship Program of China Association of Science and Technology under Grant YESS20220372.

\bibliographystyle{elsarticle-num-names} 
\bibliography{references}
\end{document}